\documentclass[times,twocolumn,final,5p]{elsarticle}
\usepackage{framed,multirow}

\usepackage{amssymb}
\usepackage{latexsym}

\usepackage{url}
\usepackage{xcolor}

\usepackage{hyperref}
\usepackage{bm}
\usepackage{url}
\usepackage{multirow}
\usepackage{ulem}
\usepackage{color}
\usepackage{xcolor}
\usepackage{amsmath,amssymb,amsfonts}
\usepackage{algorithmic}
\usepackage{graphicx}
\usepackage{textcomp}
\usepackage{bm}
\usepackage{url}  
\usepackage{multirow}
\usepackage{ulem}
\usepackage{color}
\usepackage{pifont}
\usepackage{balance}
\usepackage{colortbl}
 \usepackage{booktabs} 
\usepackage{tabularx} 
\usepackage[table]{xcolor}
\usepackage{threeparttable}
\usepackage{placeins}
\usepackage{afterpage}
\usepackage{caption}

\definecolor{newcolor}{rgb}{.8,.349,.1}

\begin{document}
	
	%\verso{Shuwei Shao \textit{et~al.}}
	
	\begin{frontmatter}
		\title{4D Monocular Surgical Reconstruction under Arbitrary Camera Motions %\tnoteref{tnote1}}%
		}
		\author[1,2,3]{Jiwei Shan}
		\cortext[cor1]{Corresponding authors}

		% \fntext[fn1]{This is author footnote for second author.
		\author[4]{Zeyu Cai}
		\author[5]{Cheng-Tai Hsieh}
            \author[1]{Yirui Li}
		% \author[1]{Wenxuan Xie}
            \author[2,3]{Hao Liu}
            \author[5]{Lijun Han$^{*}$}
            \ead{lijun_han@sjtu.edu.cn}
            \author[5,6]{Hesheng Wang}
		\author[1]{Shing Shin Cheng$^{*}$}
		\ead{sscheng@cuhk.edu.hk}
		
		\address[1]{Department of Mechanical and Automation Engineering and T Stone Robotics Institute, The Chinese University of Hong Kong, Hong Kong.}
        \address[2]{Shenyang Institute of Automation, Chinese Academy of Sciences, Shenyang, China}
        \address[3]{State Key Laboratory of Robotics and Intelligent Systems, Shenyang, China}
		\address[4]{School of Integrated Circuits, Shanghai Jiao Tong University, Shanghai, China}
		\address[5]{School of Automation and Intelligent Sensing, Shanghai Jiao Tong University, Shanghai, China}
            \address[6]{Key Laboratory of System Control and Information Processing, Ministry of Education of China, Shanghai, China}
		
		\normalem
	
		\begin{abstract}
Reconstructing deformable surgical scenes from endoscopic videos is a challenging task with important clinical applications. Recent state-of-the-art approaches, such as those based on implicit neural representations or 3D Gaussian splatting, have made notable progress in this area. However, most existing methods are designed for deformable scenes with fixed endoscope viewpoints and rely on stereo depth priors or accurate structure-from-motion for both initialization and optimization. This limits their ability to handle monocular sequences with large camera movements, restricting their use in real clinical settings. To address these limitations, we propose Local-EndoGS, a high-quality 4D reconstruction framework for monocular endoscopic sequences with arbitrary camera motion. Local-EndoGS introduces a progressive, window-based global scene representation that allocates local deformable scene representations for each observed window, enabling scalability to long sequences with substantial camera movement. To overcome unreliable initialization due to the lack of stereo depth or accurate structure-from-motion, we propose a coarse-to-fine initialization strategy that integrates multi-view geometry, cross-window information, and monocular depth priors, providing a robust foundation for subsequent optimization. In addition, we incorporate long-range 2D pixel trajectory constraints and physical motion priors to improve the physical plausibility of the recovered deformations. We comprehensively evaluate Local-EndoGS on three public endoscopic datasets with deformable scenes and varying camera motions. Local-EndoGS achieves superior performance in both appearance quality and geometry, consistently outperforming state-of-the-art methods. Extensive ablation studies further validate the effectiveness of our key designs. Our code will be released upon acceptance at \url{https://github.com/IRMVLab/Local-EndoGS}.
		\end{abstract}
		
		
	\end{frontmatter}
	
	%\linenumbers
	
	%% main text
\section{Introduction}\label{sec:introduction}
	
Endoscopes are widely used to examine almost all anatomical structures in the human body. Many diseases can also be treated with endoscopes equipped with specialized medical instruments. As endoscopic imaging and computational technologies advance, high-quality surgical reconstruction from endoscopic images is becoming increasingly important in medical applications. For example, it enables virtual and augmented reality tools for surgical simulation and training, which help improve learning and practical skills~\cite{ota1995virtual}. High-quality reconstruction also enhances visualization for diagnosis and supports accurate preoperative planning. Detailed 3D models of patient-specific anatomy help surgeons understand complex structures and reduce surgical risks~\cite{maier2017surgical}. However, achieving high-quality surgical reconstruction presents several challenges. First, physiological movements such as breathing and heartbeat, as well as interactions between surgical instruments and soft tissue, cause the surgical scene to deform. In addition, the confined space within the body limits the size of endoscopes, which prevents direct acquisition of depth information. This restriction also limits possible viewing angles, resulting in fewer three-dimensional cues from the surgical scene. To address these challenges, various reconstruction algorithms have been proposed. These include traditional methods based on depth estimation or SLAM~\cite{schmidt2024tracking}, methods using implicit neural representations~\cite{endonerf,endosurf,forplane,li2024sdfplane,shan2025uw,lightneus,enerf-slam,uc-nerf,dds-slam}, and methods based on 3D Gaussian splatting~\cite{3dgs,yang2024deform3dgs,eh-surgs,li2024endosparse,xie2024surgicalgaussian,liu2024lgs,huang2024endo,liu2025foundation,free-surgs,endogslam}.

\begin{figure}[t]
\centerline{\includegraphics[width=1\linewidth]{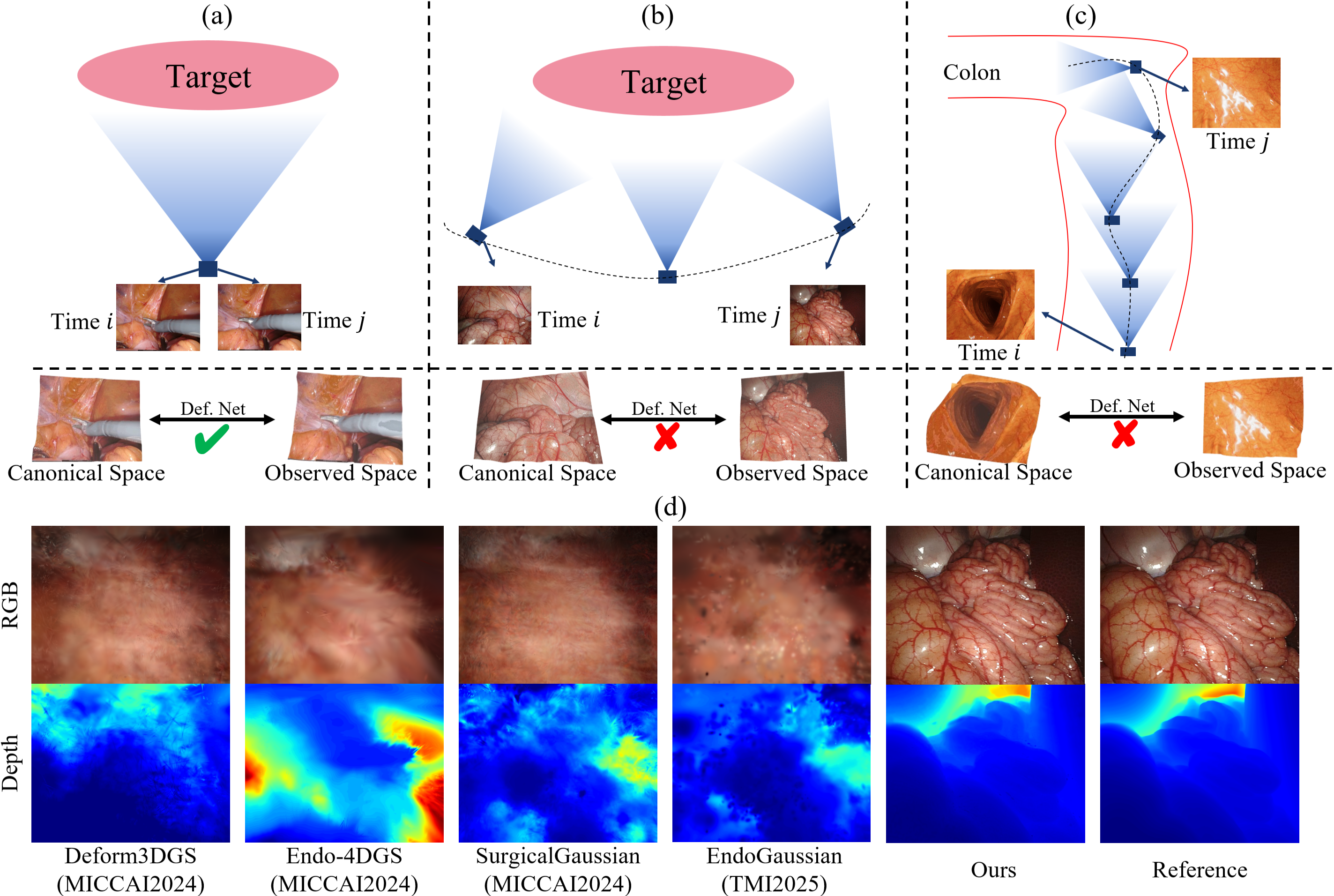}}
\caption{
(a)--(c) Illustrations of three typical types of camera motion in surgical scenes: (a) fixed camera, (b) camera moving around the tissue, and (c) camera moving forward. (d) Monocular reconstruction results of different methods under camera motion. State-of-the-art 4D surgical reconstruction algorithms experience significant degradation in reconstruction quality when the camera moves, while our method maintains superior performance.
}
    \label{fig:fig1}
\end{figure}

Traditional surgical scene reconstruction algorithms have been extensively developed over the past decades and have achieved significant progress~\cite{schmidt2024tracking}. However, these methods either assume static conditions or are unable to effectively capture the complex topological changes associated with soft tissue deformation~\cite{endonerf}. In recent years, implicit neural methods, which use neural networks to represent three-dimensional spaces implicitly, have shown improved performance in several tasks~\cite{tewari2020state,tewari2022advances}, including surgical scene reconstruction. Most INR methods model deformable surgical scenes using a single canonical space and a deformation field, as illustrated in Fig.~\ref{fig:fig1}. The canonical space typically represents the three-dimensional state of the scene at a reference time (usually set as \(t = 0\)) and is implicitly modeled by a multilayer perceptron (MLP). The deformation network establishes correspondences between the canonical space and the observed space (\(t > 0\)), thereby capturing scene deformations. During training, the model parameters—including both the canonical space MLP and the deformation network—are optimized by minimizing the loss between rendered results and the input RGB images, as well as stereo depth priors~\cite{endonerf,endosurf,forplane,li2024sdfplane,shan2025uw,enerf-slam,dds-slam}. Compared with traditional algorithms, INR methods can provide high-quality reconstructions of deformable scenes and enable photorealistic rendering. However, both during training and inference, INR methods require sampling a large number of rays and points in three-dimensional space, with neural network inference at each sample. This greatly increases computational cost and leads to longer training times and slower inference speeds~\cite{tewari2020state,tewari2022advances}. Although there are many works aimed at accelerating training and inference~\cite{muller2022instant,fridovich2023k}, it remains challenging to meet the practical requirements of medical scenarios.

Recently, 3D Gaussian Splatting (3DGS)~\cite{3dgs} has emerged as a promising technique for novel view synthesis. It can achieve performance comparable to or better than state-of-the-art INR methods, while reducing training time and greatly improving rendering speed. Building on this progress, many methods~\cite{yang2024deform3dgs,eh-surgs,li2024endosparse,xie2024surgicalgaussian,liu2024lgs,huang2024endo,liu2025foundation} have extended 3DGS to 4D surgical reconstruction. These methods model deformable scenes using a single canonical space and a deformation network, similar to INR approaches. The main difference is that they use 3DGS to represent the canonical space, leveraging its strengths to improve both the quality and speed of deformable reconstruction. Despite these advances, several key challenges remain. \textbf{First}, most existing methods~\cite{yang2024deform3dgs,eh-surgs,li2024endosparse,xie2024surgicalgaussian,liu2024lgs,huang2024endo,liu2025foundation} are designed for scenarios where the endoscope remains fixed (see Fig.~\ref{fig:fig1}(a)). In these cases, a single canonical space and a single deformation network can establish correspondences between the observed space and the canonical space, effectively capturing scene deformations. However, when the endoscope moves significantly (see Fig.~\ref{fig:fig1}(b) and Fig.~\ref{fig:fig1}(c)), new scene content constantly enters the field of view, and the observed scene may be completely different from what is already represented in the canonical space. As a result, these methods struggle to associate new observations with the canonical space. \textbf{Second}, to obtain a good initialization, current state-of-the-art algorithms usually rely on stereo depth priors or structure-from-motion (SfM) algorithms such as COLMAP~\cite{schonberger2016structure} to initialize the canonical space. However, for monocular endoscopes, depth priors often have scale ambiguity. Moreover, although COLMAP performs well for static natural scenes, it has difficulty with deformable endoscopic scenes, especially under challenging conditions such as lighting changes and limited texture. Because of these limitations, current state-of-the-art methods often show a significant drop in reconstruction quality, or even fail completely, when applied to monocular surgical reconstruction with arbitrary camera motions (as shown in Fig.~\ref{fig:fig1}(d)).

To address these challenges, we propose \textbf{Local-EndoGS}, a high-quality 4D surgical reconstruction framework for monocular endoscopic sequences with arbitrary camera motion. Similar to state-of-the-art algorithms, our framework builds on 3D Gaussian Splatting~\cite{3dgs} and leverages its efficient rendering capabilities. To effectively capture dynamic changes in long surgical sequences with significant camera motion, we design a progressive, window-based global scene representation to model deformable surgical scenes. Specifically, we dynamically divide the input sequence into multiple local windows based on the scene’s dynamics and use local deformable scene representations to model the content observed in each window. Each local deformable scene representation consists of a local canonical space and a deformation field, with parameters optimized progressively. This approach ensures scalability and allows the framework to handle long sequences with significant camera motion. To overcome unreliable initialization due to the lack of stereo depth or accurate SfM in monocular endoscopic sequences, we introduce a coarse-to-fine initialization strategy for the local canonical space in each window. This strategy integrates multi-view geometric information, cross-window information, and monocular depth priors to provide stable initialization and maintain scale consistency, eliminating the need for stereo depth priors and COLMAP. Furthermore, we incorporate long-range 2D pixel trajectory constraints and physical motion priors into our optimization framework to ensure that the recovered deformations accurately reflect real tissue motion. As shown in Fig.~\ref{fig:fig1}(d), experiments on public datasets demonstrate that our method outperforms existing state-of-the-art methods in reconstructing deformable scenes from monocular endoscopy under large camera motions.

% This capability is important for medical applications, such as creating patient-specific virtual anatomical models and providing surgical simulations for clinical training. 

In summary, our main contributions are as follows:
\begin{itemize}
    \item \textbf{A scalable 4D reconstruction framework for monocular endoscopy:} To the best of our knowledge, Local-EndoGS is the first framework that enables high-quality 4D reconstruction of deformable surgical scenes from monocular endoscopic sequences with arbitrary camera motion. Our method uses 3D Gaussian Splatting and a progressive, window-based global scene representation to efficiently and accurately model long sequences.
    \item \textbf{Coarse-to-fine initialization strategy for monocular sequences:} We introduce a robust coarse-to-fine initialization strategy for local canonical spaces. This strategy integrates multi-view geometry, cross-window information, and monocular depth priors. It removes the need for stereo depth or accurate structure-from-motion, ensuring stable and scale-consistent initialization for monocular sequences.
    \item \textbf{Incorporation of long-range trajectory and motion priors in optimization:} We incorporate long-range 2D pixel trajectory constraints and physical motion priors into our optimization framework. This enables more accurate and robust deformation estimation when reconstructing dynamic surgical scenes from monocular sequences.
    \item \textbf{Comprehensive evaluation and demonstration of effectiveness:} We conduct rigorous evaluations on multiple datasets and perform thorough ablation studies to demonstrate the effectiveness and advantages of Local-EndoGS.
\end{itemize}
	
\section{Related work}
In this work, we focus on 4D monocular surgical reconstruction under arbitrary camera motion. Thus, we primarily review the following areas: 1) surgical reconstruction methods based on non-implicit neural representations; 2) surgical reconstruction methods based on implicit neural representations; and 3) techniques utilizing 3D Gaussian Splatting.

\subsection{Surgical Reconstruction Based on Non-Implicit Neural Representations}
In recent decades, notable progress has been made in reconstructing surgical scenes using non-implicit methods. These techniques are widely used in clinical fields such as orthopedics, otolaryngology, gastroenterology, and pulmonology \cite{schmidt2024tracking,cui2021virtual,xie2022mixed}. Early methods often integrated Simultaneous Localization and Mapping (SLAM) with densification techniques to generate semi-dense or dense representations of surgical scenes~\cite{mahmoud2017orbslam,mahmoud2018live,chen2018slam,marmol2019dense}. However, since these methods assume that the scene remains static, their performance is limited when applied to deformable environments. To overcome these limitations, researchers have introduced algorithms tailored for non-rigid scenarios, such as Non-Rigid Structure from Motion (NRSfM)~\cite{sengupta2021colonoscopic,lamarca2020defslam} and Shape-from-Template (SfT) approaches~\cite{cheema2018image,lamarca2020defslam}. NRSfM methods frequently incorporate assumptions about tissue motion, including low-rank shape models or isometric constraints, to make the problem more tractable. SfT approaches first establish a scene template or utilize a predefined template (for instance, sheet-like or tubular structures), and subsequently align this template to each frame to monitor deformations. With the development of deep learning, recent research has explored learning-based stereo depth estimation and deformation tracking using sparse deformation fields~\cite{li2020super,long2021dssr}. Although these methods enhance reconstruction flexibility and accuracy, they continue to encounter difficulties in managing topological variations and achieving photorealistic rendering results.

\subsection{Surgical Reconstruction Based on Implicit Neural Representations}

Implicit neural representations have emerged as a promising technology in recent years, addressing the limitations of traditional surgical reconstruction algorithms through differentiable rendering and neural networks~\cite{endonerf,endosurf,forplane}. EndoNeRF~\cite{endonerf} was an early attempt to use implicit neural representations for modeling deformable surgical scenes. It employs a canonical neural radiance field and a time-varying deformation network to simulate tissue deformation, achieving promising results. EndoSurf~\cite{endosurf} builds on this approach by using signed distance fields to represent scene geometry. Lerplane, Forplane~\cite{forplane}, and SDFPlane~\cite{li2024sdfplane} improve training efficiency by decomposing four-dimensional space into several orthogonal two-dimensional feature planes. UW-DNeRF~\cite{shan2025uw} further incorporates uncertainty in depth priors and leverages local information. Despite these advances, most methods assume a stationary endoscope and typically rely on stereo depth priors. Several algorithms are capable of handling scenarios involving endoscope motion. LightNeuS~\cite{lightneus} reconstructs static scenes from monocular endoscopic images while accounting for lighting changes. ENeRF-SLAM~\cite{enerf-slam} and DDS-SLAM~\cite{dds-slam} use implicit neural representations to build dense SLAM systems for endoscopy. UC-NeRF~\cite{uc-nerf} employs an uncertainty-aware conditional NeRF for novel view synthesis in endoscopic scenes. However, most of these methods are designed for static scenes and do not consider scene deformation. In addition, they face challenges such as long training times and low rendering efficiency due to the computational complexity of implicit neural representations.

\begin{figure*}[t]
\centerline{\includegraphics[width=0.99\linewidth]{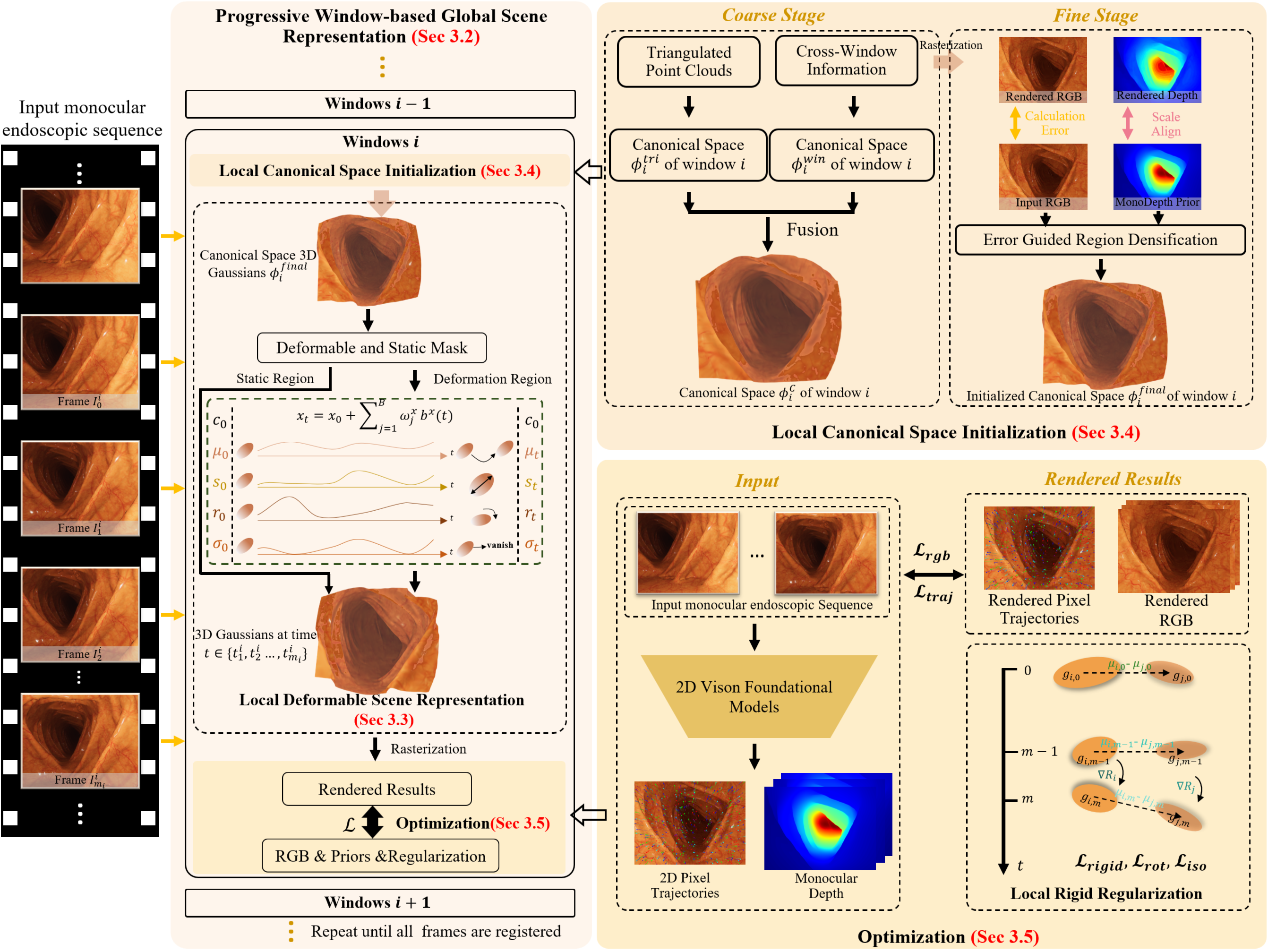}}
    % \vspace{-6pt}
\caption{Overview of Local-EndoGS. Given a long monocular endoscopic sequence with arbitrary camera motion, Local-EndoGS reconstructs the entire deformable scene using a progressive window-based global scene representation (\ref{sec:Progressive}). Specifically, the sequence is first divided into multiple local windows based on its dynamic characteristics. For each local window, the scene structure is initialized using a local canonical space initialization strategy (\ref{sec:init}). Then, a local deformable scene representation (\ref{sec:local}) is used to model each region. The parameters of each local scene representation are then optimized using carefully designed loss functions (\ref{sec:loss}) in a progressive manner until all local models are fully optimized.}
    % \vspace{-18pt}
    \label{fig:overview}
\end{figure*}

\subsection{Surgical Reconstruction Based on 3D Gaussian Splatting}

3D Gaussian Splatting (3DGS), introduced by~\cite{3dgs}, is an explicit radiance field method that enables efficient and high-quality rendering of 3D scenes. To further improve training and rendering efficiency, many approaches utilize 3DGS in the canonical space and combine it with a deformation field to model deformable surgical scenes. The deformation field can be represented in various ways. For example, Deform3DGS~\cite{yang2024deform3dgs} and EH-SurGS~\cite{eh-surgs} use explicit basis functions, while EndoSparse~\cite{li2024endosparse} and SurgicalGaussian~\cite{xie2024surgicalgaussian} employ multilayer perceptrons (MLPs) to capture deformation. LGS~\cite{liu2024lgs}, Endo-4DGS~\cite{huang2024endo}, and EndoGaussian~\cite{liu2025foundation} combine multiple orthogonal 2D feature planes with a compact MLP. Compared to implicit neural representation methods, these approaches significantly improve training speed and rendering efficiency. However, they share similar limitations with implicit methods: they assume a stationary endoscope and typically rely on stereo depth priors or precise structure-from-motion for initialization and optimization, which limits their applicability in real surgical scenarios. Free-SurGS~\cite{free-surgs}, EndoGSLAM~\cite{endogslam} {\color{black}and Endoflow-SLAM \cite{wu2025endoflow}} use 3DGS for surgical scene reconstruction and joint camera pose optimization. {\color{black}Endo-2DTAM~\cite{huang2025advancing} introduces a surface-normal-aware pipeline based on 2D Gaussian distributions~\cite{huang20242d}. {Gaussian Pancakes}~\cite{bonilla2024gaussian} combined 3D Gaussian Splatting with RNN-SLAM \cite{ma2021rnnslam} to enable real-time, high-quality 3D reconstruction of endoscopic scenes, offering substantial gains in both rendering accuracy and computational efficiency.} However, these methods are designed for static scenes and do not address the deformable nature of endoscopic images. In this work, we address these limitations by proposing a 4D reconstruction algorithm that supports arbitrary camera motion and enables reconstruction from monocular endoscopic images, providing high-quality reconstruction of deformable surgical scenes.

\section{Methodology}
\label{method}

Given a sequence of monocular endoscopic images $\{I_i\}_{i=1}^Q$ with arbitrary but known poses $\{E_i=(R_i, t_i)\}_{i=1}^Q$ and intrinsic parameters $K$, our goal is to develop a 4D monocular surgical reconstruction framework based on 3DGS that accurately captures deformable scenes with fine detail and geometric structure. The Local-EndoGS pipeline is illustrated in Fig.~\ref{fig:overview} and consists of four main components:
1) A \textit{progressive window-based global scene representation} that adaptively divides the input sequence into smaller windows and progressively optimizes the parameters of each local model to represent the scene within each window (Sec.~\ref{sec:Progressive});
2) A \textit{local deformable scene representation} that models the deformable scene within each window (Sec.~\ref{sec:local});
3) A \textit{local canonical space initialization} that reliably initializes the local canonical space in each window to ensure stable model performance (Sec.~\ref{sec:init});
4) A \textit{carefully designed loss function} that optimizes the model by incorporating monocular images, long-range 2D pixel trajectory priors, and physical motion priors (Sec.~\ref{sec:loss}).
	
\subsection{Preliminaries: 3DGS-based Deformable Surgical Reconstruction}
\label{sec:3dgs&def3dgs}

\textbf{3D Gaussian Splatting.} 3DGS~\cite{3dgs} represents 3D scenes explicitly using anisotropic 3D Gaussian functions \( \phi = \{g_i(x)\} \). Each Gaussian \( g_i \) is defined by its position \( \mu \in \mathbb{R}^3 \), covariance matrix \( \Sigma \in \mathbb{R}^{3 \times 3} \), opacity \( \sigma \in \mathbb{R} \), and view-dependent color \( c \in \mathbb{R}^3 \), which is parameterized by spherical harmonics~\cite{fridovich2022plenoxels}. The distribution of a 3D Gaussian is given by:
\begin{equation}
    g_i(x) = e^{-\frac{1}{2} (x-\mu)^T \Sigma^{-1} (x-\mu)}
    \label{equ:3dgs}
\end{equation}
To ensure that \( \Sigma \) remains a valid covariance matrix during optimization, it is parameterized as \(\Sigma = R S S^T R^T\), where \( S \) is a diagonal scale matrix and \( R \) is a rotation matrix, defined by a scaling vector \( s \in \mathbb{R}^3 \) and a quaternion \( r \in \mathbb{R}^4 \).

To render an image from a given viewpoint, the 3D Gaussians are projected onto the 2D image plane using splatting techniques~\cite{ewa}. The corresponding 2D covariance matrix \( \Sigma^{\prime} \) and center \( \mu^{\prime} \) in camera coordinates are computed as
$\Sigma^{\prime} = J W \Sigma W^T J^T$ and $\mu^{\prime} = J W \mu$, where \( W \) denotes the viewing transformation and \( J \) is the Jacobian of the affine approximation of the projective transformation. For each pixel \( p \), the color \( \hat{C}(p) \) is computed by alpha blending:
\begin{equation}
\hat{C}(p) = \sum_{i \in N} c_i \alpha_i \prod_{j=1}^{i-1}\left(1 - \alpha_j\right)
\label{equ:rgb}
\end{equation}
Here, \( N \) is the set of all Gaussians that influence pixel \( p \) along the viewing ray, and the final opacity \( \alpha_i \) is given by the product of the learned opacity \( \sigma_i \) and the Gaussian:
\begin{equation}
\alpha_i = \sigma_i \exp\left(-\frac{1}{2}(p - \mu_i^{\prime})^T {\Sigma^{\prime}}^{-1} (p - \mu_i^{\prime})\right)
\end{equation}
where \( \mu_i^{\prime} \) is the projected 2D center of the \( i \)-th Gaussian in the camera coordinate system. Similarly, the depth \( \hat{D}(p) \) is rendered as:
\begin{equation}
\hat{D}(p) = \sum_{i \in N} d_i \alpha_i \prod_{j=1}^{i-1}\left(1 - \alpha_j\right)
\label{equ:depth}
\end{equation}
Here, \( d_i \) is the z-depth coordinate of the \( i \)-th 3D Gaussian in view space.

\textbf{3DGS-based Deformable Surgical Reconstruction.}
Recent algorithms for deformable surgical scene reconstruction~\cite{yang2024deform3dgs,eh-surgs,li2024endosparse,xie2024surgicalgaussian,liu2024lgs,huang2024endo,liu2025foundation} typically use a single canonical space, defined at a reference time (often \( t = 0 \)), and represented by 3D Gaussians. To model tissue motion and deformation during surgery, a deformation network \( D^t \) predicts changes in the center position, rotation, and scale for each Gaussian at each time step \( t \):
\begin{equation}
(\Delta\mu_t,\, \Delta r_t,\, \Delta s_t) = D^t(\mu_0,\, r_0,\, s_0, t)
\label{equ:delta}
\end{equation}
Here, \( \mu_0, r_0, s_0 \) are the center coordinates, rotation quaternion, and scaling factors in the canonical space. $\Delta\mu_t,\, \Delta r_t,\, \Delta s_t$ are the temporal offsets at time \( t \) relative to their canonical values. The parameters at time \( t \) are obtained by adding these offsets to the canonical values:
\begin{equation}
(\mu_t,\, r_t,\, s_t) = (\mu_0 + \Delta\mu_t,\, r_0 + \Delta r_t,\, s_0 + \Delta s_t)
\label{equ:delta+}
\end{equation}
Finally, during training, RGB and depth images are rendered using \eqref{equ:rgb} and \eqref{equ:depth}, respectively. The ground truth RGB images are provided as inputs, while depth priors are obtained from a stereo depth estimation network~\cite{li2021revisiting}. The RGB and depth losses are computed by comparing the rendered images with the ground truth RGB images and estimated depth priors, respectively. The parameters of both the deformation network and the 3D Gaussians are jointly optimized by minimizing the combined loss.

\subsection{Progressive Window-based Global Scene Representation}
\label{sec:Progressive}

Modeling a long sequence with a single deformable 3DGS representation is challenging, especially when significant camera motion occurs. Large camera movements disrupt the correspondence between the observed and canonical spaces, often resulting in reconstruction failure. To address this limitation, we propose a progressive, window-based global scene representation.

As shown in Fig.~\ref{fig:overview}, our method first uses adaptive window partitioning to divide the input sequence into $M$ contiguous local windows. Each window is modeled using a local deformable scene representation (see Sec.~\ref{sec:local}). The entire scene is then reconstructed from multiple local models, denoted as $\{\phi_i, D_i^{t}\}_{i=1}^M$. During training, the parameters of each local model are optimized progressively. Specifically, the frames within window $i-1$ are used to optimize the parameters of the corresponding local model $\{\phi_{i-1}, D_{i-1}^{t}\}$. After optimizing $\{\phi_{i-1}, D_{i-1}^{t}\}$, we save its parameters and move on to optimize those for window $i$. This process is repeated until all $M$ local models have been optimized. The progressive optimization strategy enables our method to efficiently handle sequences of arbitrary length and various types of camera motion.

To balance model accuracy and training efficiency, we propose an adaptive method that determines window sizes based on the dynamic characteristics of the input sequence. We analyze sequence dynamics from two perspectives: camera motion and frame content variation. For camera motion, we measure both translational and rotational pose changes throughout the {\color{black}ground-truth camera poses} of each sequence. A new window is created when either the translational difference exceeds the threshold {\color{black}$\delta_t$} or the rotational angle difference exceeds the threshold {\color{black}$\delta_r$}. This approach ensures consistent viewpoints within each window and reduces representation errors caused by large viewpoint changes. For frame content variation, previous methods~\cite{shaw2024swings} rely on optical flow to estimate scene changes. However, optical flow is often unreliable in endoscopic environments due to illumination changes and textureless tissue surfaces. Instead, we use a simple yet effective method by comparing RGB differences between each frame and the first frame of the current window. A new window is created when this difference exceeds a predefined threshold (empirically set to 0.05). Compared to using a fixed window size as a hyperparameter, our adaptive partitioning strategy increases the sampling frequency in highly dynamic regions and reduces it in low-motion areas. This approach leads to improved system performance, as demonstrated in the experimental results (see Sec.~\ref{sec:ablation}).

\subsection{Local Deformable Scene Representation}
\label{sec:local}

Dividing the input sequence into multiple local windows with similar content (as described in Sec.~\ref{sec:Progressive}) ensures that the content within each window can be effectively modeled using a single canonical space and a single deformation network. Therefore, any 3DGS-based deformable reconstruction method can be applied to each window. In this work, we use EH-SurGS~\cite{eh-surgs} for this purpose.

Specifically, for each local window, we construct a canonical space using 3D Gaussians (initialized as described in Sec.~\ref{sec:init}) and model local deformations using a deformation network. The deformation network predicts the temporal evolution of both the spatial parameters and the opacity of each Gaussian, which are defined as learnable functions of time. This design allows the model to flexibly capture both general and irreversible deformations:
\begin{equation}
x_t = x_0 + \sum_{j=1}^{B} \omega_j^{x} b^{x}(t).
\label{equ:x}
\end{equation}
Here, $x_t$ denotes the mean, rotation, scale, or opacity; $x_0$ is the canonical value; $b(t)$ is a Gaussian basis function with learnable center and variance; and $\omega_j$ are learnable weights.

{\color{black}To handle irreversible and dynamic scene changes caused by intraoperative operations such as tissue shearing, EH-SurGS introduces a {life-cycle mechanism} for 3D Gaussians. Each Gaussian is activated only within its valid temporal range and deactivated once the corresponding structure disappears.}
To improve computational efficiency, EH-SurGS uses an adaptive motion hierarchy strategy to distinguish between deformable and static regions in each local scene. A dynamically updated mask separates these regions based on average deformation and the consistency of rendering loss. This approach allows for efficient resource allocation during training and inference. More details are provided in our previous work~\cite{eh-surgs}.

\subsection{Local Canonical Space Initialization}
\label{sec:init}

A well-initialized 3D Gaussian representation in the canonical space is important for effective model optimization and performance. Previous methods often use stereo depth priors or structure-from-motion (SfM) point clouds for scene initialization~\cite{yang2024deform3dgs,eh-surgs,li2024endosparse,xie2024surgicalgaussian,liu2024lgs,huang2024endo,liu2025foundation}. However, for monocular sequences, scale ambiguity in depth estimation makes initialization challenging. In addition, unique characteristics of endoscopic scenes often result in sparse and unstable SfM point clouds. To address these issues, we propose a coarse-to-fine initialization method for monocular sequences. Our approach combines multi-view geometry, cross-window information, and monocular depth priors to achieve stable and consistent initialization.

\begin{figure}[t]
    \centering
\includegraphics[width=0.45\textwidth]{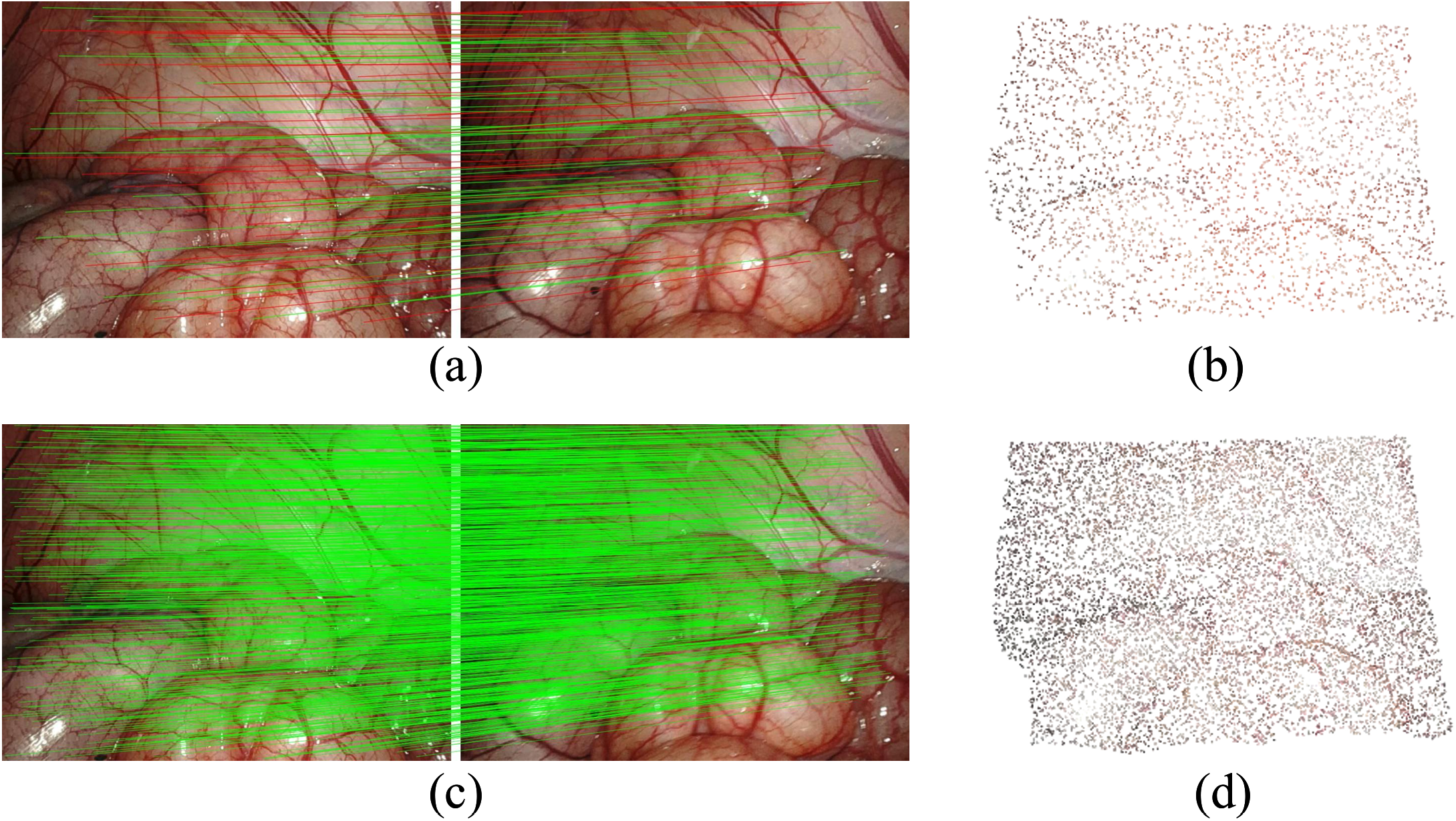}
    \caption{
Comparison of Feature Matching and Point Cloud Results Using Traditional Methods and Track-Any-Point (TAP) model \cite{chen2024leap}.
{\color{black}(a) Correspondences obtained using SIFT keypoints with brute-force matching.
(b) Sparse point cloud reconstructed from  the correspondences shown in (a).}
(c) Correspondences from the TAP model.
(d) Dense point cloud from TAP-based correspondences.
Green and red lines indicate correct and incorrect feature matches, respectively.
    }
\label{fig:ablation_tap_visualization}
\end{figure}

\subsubsection{Coarse Stage: Scale-Aware Initialization}

In this stage, we construct a dense point cloud to initialize 3D Gaussians in the local canonical space. This point cloud captures the basic geometry and maintains a consistent global scale, providing a stable basis for the next stage.

{\color{black}For each local window $i$, composed of an image set $S_i = \{I_j^i\}_{j=0}^{m_i}$ with timestamps $\{t_j^i\}_{j=0}^{m_i}$, we use the known ground-truth poses of all frames to generate a dense triangulated point cloud}. {\color{black}A common traditional approach is to use traditional feature extraction and matching methods, such as SIFT keypoints \cite{lowe2004distinctive} with brute-force matching in OpenCV, followed by multi-view triangulation to obtain an initial point cloud.} However, as shown in Fig.~\ref{fig:ablation_tap_visualization}a, challenges such as illumination changes, repetitive low-texture patterns, and tissue deformation in endoscopic scenes can lead to incorrect feature matching and sparse feature points. As a result, the generated point cloud is often sparse (Fig.~\ref{fig:ablation_tap_visualization}b). To address this, {\color{black}we leverage the Track-Any-Point (TAP) model~\cite{chen2024leap}, a 2D vision foundation model built upon the CoTracker framework~\cite{karaev2024cotracker}, to perform end-to-end point tracking for establishing correspondences across multiple image frames.} Specifically, we use this pre-trained TAP model to extract pixel-wise temporal trajectories for $K$ sampled pixels. These trajectories are denoted as $\mathcal{T}_i = \{U(p_q)(t_j^i) \mid q=1,\ldots,K;~ j=0,\ldots,m_i\}$, where $p_q \in \mathbb{R}^2$ is the location of the $q$-th pixel in $I_0^i$, and $U(p_q)(t_j^i)$ is its 2D position in the $j$-th frame of window $i$. We then select $k$ image pairs from window $i$ and use $\mathcal{T}_i$ to find correspondences across frames, as shown in Fig.~\ref{fig:ablation_tap_visualization}b. These correspondences are triangulated using known camera intrinsics and poses to recover 3D points. The point clouds from multiple pairs are merged to obtain a scale-consistent dense point cloud $P_i^{tri}$. As shown in Fig.~\ref{fig:ablation_tap_visualization}d, the point cloud obtained using our TAP-based method is much denser than that produced by traditional methods and better represents the geometric structure of the scene. This point cloud is used to initialize the 3D Gaussians $\phi_i^{tri}$, following~\cite{3dgs}, and serves as a global scale reference for later stages.

In addition, we introduce a cross-window information propagation strategy. With our progressive window-based global scene representation, we can efficiently propagate and integrate information across adjacent windows. The core idea is to use the optimized local deformable scene representation from window \( i-1 \) to estimate the initial canonical space in window \( i \), transferring this prior knowledge forward. Specifically, the optimized local window \( i-1 \) is represented by \( \phi_{i-1} \) and its deformation network \( D_{i-1}^t \). We apply \( D_{i-1}^t \) to the 3D Gaussians in \( \phi_{i-1} \) to predict their parameters at the observation time \( t_0^i \) in the canonical space of window \( i \), as shown in Eq.~\eqref{equ:delta} and Eq.~\eqref{equ:delta+}. This process produces a set of 3D Gaussians for window \( i \), denoted as \( \phi_i^{win} \). We then fuse \( \phi_i^{win} \) with the triangulated representation \( \phi_i^{tri} \) to form the initial representation \( \phi_i^{C} \) for the coarse stage. For the first window, we use only the triangulated point cloud for coarse stage initialization.

\begin{figure}[t]
\centerline{\includegraphics[width=1\linewidth]{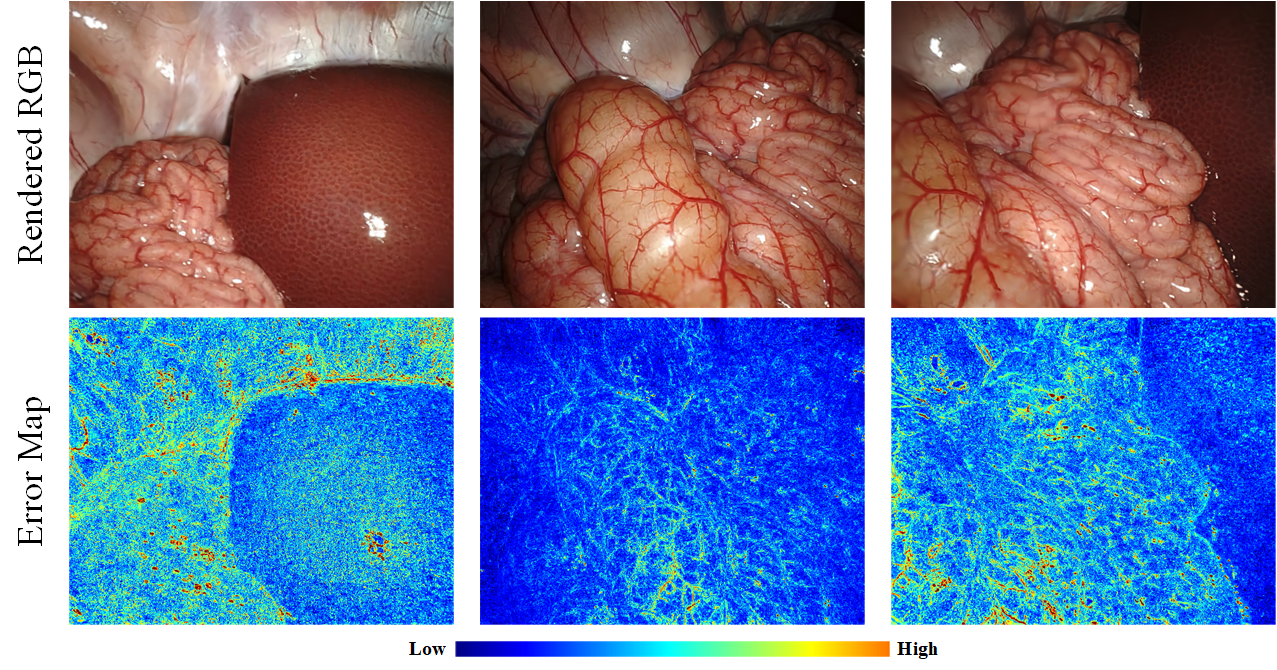}}
    % \vspace{-7pt}
\caption{
    Visualization of RGB images rendered by 3DGS $\phi^C_i$ initialized from the coarse stage and their corresponding reconstruction error maps with respect to the ground truth. 
    \textbf{Top row}: rendered RGB images. \textbf{Bottom row}: pixel-wise reconstruction error maps, where higher values indicate greater reconstruction errors, especially near tissue boundaries and regions with deformation.
}
    % \vspace{-19pt}
    \label{fig:error_map}
\end{figure}

\subsubsection{Fine Stage: Error-Guided Region Refinement}

As shown in Fig.~\ref{fig:error_map}, we observe noticeable reconstruction errors in some regions, especially near tissue boundaries, specular reflections, and areas with significant deformation. To address this, we propose an error-guided region refinement strategy that incorporates monocular depth priors. Specifically, we first render the RGB image and the corresponding depth map for the first frame \( I_0^i \) of window \( i \) using the current initialization \( \phi_i^{C} \), denoted as \( I_{\text{render}} \) and \( D_{\text{render}} \), according to Eq.~\eqref{equ:rgb} and Eq.~\eqref{equ:depth}, respectively. At the same time, we use a pretrained monocular depth estimation network~\cite{yang2024depth} to predict a depth map \( D_{\text{est}} \) from \( I_0^i \).  We find that $D_{\text{render}}$ benefits from strict geometric consistency with the reconstructed 3D scene, but may be inaccurate or incomplete in regions affected by erroneous 3D initialization. In contrast, $D_{\text{est}}$ can provide consistent depth estimates even in challenging regions, but suffers from scale ambiguity. Therefore, we align the monocular depth with the rendered depth by estimating a scale factor \( \alpha \) and an offset \( \beta \) using least-squares fitting:
\begingroup
\begin{equation}
    D_{\text{fine}} = \alpha \cdot D_{\text{est}} + \beta.
    \label{equ:depth1}
\end{equation}
\endgroup
where
\begin{equation}
    \alpha, \beta = \arg\min_{\alpha, \beta} \| D_{\text{render}} - (\alpha D_{\text{est}} + \beta) \|_2^2.
    \label{equ:depth2}
\end{equation}
$D_{\text{fine}}$ combines the strengths of both sources and mitigates their individual limitations.

Next, we compute a per-pixel photometric error map by comparing \( I_{\text{render}} \) with the observed image \( I_0^i \). For regions with low error (empirically set to 0.7), we retain the initialization from the coarse stage. For regions with high error, we use the aligned depth map \( D_{\text{fine}} \) to back-project the corresponding pixels into 3D space. We then generate new 3D points that are fused with the original 3D Gaussians to refine the geometry. This selective refinement is both computationally efficient and helps preserve well-initialized geometry in areas with low error.

In summary, our initialization strategy first uses multi-view geometry and cross-window information to address scale ambiguity in monocular depth estimation. In cases where the camera remains stationary and multi-view geometry becomes ineffective, we utilize cross-window information to maintain scale consistency among local windows. Monocular depth priors are then applied in the fine stage to refine regions where the coarse initialization is less accurate. This approach is specifically designed for monocular sequences and does not require stereo depth information. As demonstrated by our experimental results (see Sec.~\ref{sec:ablation}), this improves the robustness and accuracy of the overall initialization.

\begin{table*}[ht]
\centering
\caption{Summary of the datasets and sequences used in our study. Each sequence's number of frames, image resolution, camera motion type, and window count are listed.}
\label{tab:dataset_summary}
\begin{tabular}{llccccl}
\toprule
\textbf{Dataset} & \textbf{Sequence} & \textbf{Frames} & \textbf{Resolution} & \textbf{Camera Motion} & \textbf{Windows} \\
\midrule
\multirow{2}{*}{{EndoNeRF~\cite{endonerf}}} 
  & Pulling  & 63  & $512 \times 640$ & Fixed camera & 1 \\
  & Cutting  & 156 & $512 \times 640$ & Fixed camera & 1 \\
\midrule
\multirow{2}{*}{{StereMIS~\cite{stereomis}}} 
  & Sequence1 & 1000 & $512 \times 640$ & Moving around tissue & 49 \\
  & Sequence2 & 1500 & $512 \times 640$ & Moving around tissue & 69 \\
\midrule
\multirow{3}{*}{{EndoMapper~\cite{endomapper}}} 
  & Sequence1 & 267 & $512 \times 640$ & Moving forward & 15 \\
  & Sequence2 & 267 & $512 \times 640$ & Moving forward & 16 \\
  & Sequence3 & 267 & $512 \times 640$ & Moving forward & 20 \\
\bottomrule
\end{tabular}
\end{table*}

\subsection{Optimization}\label{sec:loss}
% \subsubsection{Loss Functions}
We carefully design three loss functions to optimize the local deformable representation in posed monocular endoscopic sequences.

\textbf{Rendering Loss.} The rendering loss enforces color consistency between the rendered and observed images for each frame within a window.  
During training, we render images \( \hat{I} \) using \eqref{equ:rgb} and compute the rendering loss \( \mathcal{L}_{rgb} \) as:
\begin{equation}
    \mathcal{L}_{rgb} = (1 - M) \left( (1 - \lambda) \mathcal{L}_1 + \lambda \mathcal{L}_{D-SSIM} \right).
\end{equation}
Here, \( M \) is a mask that excludes regions containing surgical instruments, as in previous work~\cite{yang2024deform3dgs,eh-surgs,li2024endosparse,xie2024surgicalgaussian,liu2024lgs,huang2024endo,liu2025foundation}. The weight \(\lambda\) is empirically set to 0.2.  
\( \mathcal{L}_1 \) is the pixel-wise \( L_1 \) loss between the rendered image \( \hat{I} \) and the input image \( I \), i.e., \( \mathcal{L}_1 = \| \hat{I} - I \|_1 \).  
\( \mathcal{L}_{D-SSIM} \) is the structural dissimilarity loss~\cite{3dgs}.  
This objective balances pixel-level accuracy and perceptual similarity, following the approach in 3D Gaussian Splatting~\cite{3dgs}. Notably, we do not use depth priors as a supervision signal, as we found that this negatively impacts the performance of our method (see Sec.~\ref{sec:mono} for details).

\textbf{2D Tracking Loss.}  
Leveraging the stable correspondences across multiple frames provided by the Track-Any-Point (TAP) model~\cite{chen2024leap}, we construct a 2D tracking loss. This loss enforces consistency between the canonical and observed spaces by supervising pixel-wise temporal trajectories in the RGB frames. Following~\cite{wang2024shape}, for each local window \(i\), we first rasterize the motion of 3D Gaussians in the canonical space at the observation time \(t_0^i\) into the query frame \(I_j^i\) at time \(t_j^i\). Specifically, we compute a 3D trajectory map \(\hat{X}_{t_0^i \to t_j^i}^w \in \mathbb{R}^{H \times W \times 3}\), which provides the 3D world coordinates at time \( t_j^i \) corresponding to each 2D pixel location initially observed at \( t_0^i \):
\begin{equation}
\hat{X}_{t_0^i \to t_j^i}^w({p}) = \sum_{i \in H({p})} \mu_{i, t_0^i \to t_j^i} \alpha_i \prod_{j=1}^{i-1}\left(1 - \alpha_j\right),
\end{equation}
where \( H({p}) \) is the set of Gaussians intersecting pixel \( p \) at \( t_0^i \), and \( \mu_{i, t_0^i \to t_j^i} \) is the center of Gaussian \( i \) at \( t_j^i \), computed from~\eqref{equ:x}. We then project these 3D points to the image plane using camera intrinsics \( K \) and extrinsics \( E_t \):
\begin{equation}
\hat{U}_{t_0^i \to t_j^i}(p) = \Pi\left( K E_t \hat{X}_{t_0^i \to t_j^i}^w(p) \right),
\end{equation}
where \( \Pi(\cdot) \) denotes the standard perspective projection. This process establishes pixel-level correspondences across frames. We supervise the rendered trajectories using 2D pixel tracks from the TAP model, as described in Sec.~\ref{sec:init}:
\begin{equation}
\mathcal{L}_{track} = \left\| U_{t_0^i \to t_j^i} - \hat{U}_{t_0^i \to t_j^i} \right\|_1.
\end{equation}

\textbf{Physics-Based Regularization.}  
To improve the physical plausibility and motion consistency of 3D Gaussian deformation modeling in local deformable scene representation, we introduce three physics-based spatial constraints to regularize the transformation of Gaussians from the canonical space to the observation space: short-term local rigidity loss (\( \mathcal{L}_{rigid} \)), local rotation similarity loss (\( \mathcal{L}_{rot} \)), and long-term local isometry loss (\( \mathcal{L}_{iso} \)). These losses are computed for each Gaussian and its \( k \) nearest neighbors (kNN) as follows~\cite{luiten2024dynamic}:
\begin{equation}
\mathcal{L}_x = \frac{1}{|\mathcal{G}|} \sum_{i \in \mathcal{G}} \sum_{j \in \mathrm{kNN}(i)} w_{i,j} \mathcal{L}_{x,i,j},
\end{equation}
where \( x \in \{rigid, rot, iso\} \), \( |\mathcal{G}| \) is the number of Gaussians in the canonical space, and \( w_{i,j} \) is an isotropic Gaussian weight that reflects the spatial relationship between the \( i \)-th and \( j \)-th Gaussians:
\begin{equation}
w_{i,j} = \exp\left(-\lambda_w \left\| {\mu}_{j,c} - {\mu}_{i,c} \right\|_2^2 \right).
\end{equation}
Here, \( \lambda_w = 2000 \) defines the standard deviation. \( {\mu}_{i,c} \) and \( {\mu}_{j,c} \) denote the positions of the \( i \)-th and \( j \)-th Gaussians in the canonical space \( c \), respectively.

The rigidity loss, \( \mathcal{L}_{{rigid}} \), encourages adjacent Gaussians within a local region to undergo similar rigid transformations. This helps preserve local structure and suppress unnatural deformations:
\begin{equation}
\mathcal{L}_{{rigid}, i,j} = \left\| ({\mu}_{j,\tau} - {\mu}_{i,\tau}) - \Delta {R}_i ({\mu}_{j,c} - {\mu}_{i,c}) \right\|_2.
\end{equation}
Here, \( \Delta {R}_i = {R}_{i,\tau} {R}_{i,c}^{-1} \) is the relative rotation of the \( i \)-th Gaussian between the canonical space \( c \) and the observation space \( \tau \).

The rotation similarity loss, \( \mathcal{L}_{{rot}} \), promotes consistent rotation among neighboring Gaussians and reduces abrupt angular changes in local regions:
\begin{equation}
\mathcal{L}_{{rot}, i,j} = \left\| \hat{q}_{j,\tau} \hat{q}_{j,c}^{-1} - \hat{q}_{i,\tau} \hat{q}_{i,c}^{-1} \right\|_2^2.
\end{equation}
Here, \( \hat{q} \) is the unit quaternion representing the rotation of each Gaussian.

Finally, the isometry loss, \( \mathcal{L}_{iso} \), preserves the relative distances between neighboring Gaussian centers over time:
\begin{equation}
\mathcal{L}_{iso, i,j} = \left\| {\mu}_{j,c} - {\mu}_{i,c} \right\|_2 - \left\| {\mu}_{j,\tau} - {\mu}_{i,\tau} \right\|_2.
\end{equation}

\textbf{Training Loss.}  
The total training loss is defined as:
\begin{equation}
\mathcal{L} = \lambda_{rgb} \mathcal{L}_{rgb} + \lambda_{track} \mathcal{L}_{track} + \lambda_{rigid} \mathcal{L}_{rigid} + \lambda_{rot} \mathcal{L}_{rot} + \lambda_{iso} \mathcal{L}_{iso}.
\end{equation}
where $\lambda_{rgb}$, $\lambda_{track}$, $\lambda_{rigid}$, $\lambda_{rot}$, and $\lambda_{iso}$ are the weights for each loss term.
In our experiments, we set $\lambda_{rgb}=1$, $\lambda_{track}=0.01$, $\lambda_{rigid}=0.05$, $\lambda_{rot}=0.05$, and $\lambda_{iso}=0.05$.

\section{Experiment Setup}
\subsection{Datasets}
We evaluate Local-EndoGS on three endoscopic datasets that feature deformable scenes and varying camera motions: the EndoNeRF dataset~\cite{endonerf}, the StereoMIS dataset~\cite{stereomis}, and the EndoMapper dataset~\cite{endomapper}. Table~\ref{tab:dataset_summary} provides a detailed summary of the dataset sequences.

\textbf{EndoNeRF dataset.} The EndoNeRF dataset~\cite{endonerf} contains six video clips recorded during a Da Vinci robotic prostatectomy procedure, with the camera remaining stationary (see Fig.~\ref{fig:fig1}(a)){\color{black}; therefore, its extrinsic parameters are fixed to the identity matrix ($R = I$, $t = 0$).} Each clip has a resolution of 512$\times$640 and a duration of 4–8 seconds at 15 fps. We use two public sequences, \textit{Pulling} and \textit{Cutting}, which depict non-rigid soft tissue deformations and contain 63 and 156 frames, respectively. We follow the protocol outlined in EndoNeRF~\cite{endonerf} for handling surgical tool occlusions. Only left-view images are used for training and evaluation.

\textbf{StereoMIS dataset.} The StereoMIS dataset~\cite{stereomis} is an in vivo collection recorded with the da Vinci Xi surgical robot. Ground-truth camera poses are obtained from the endoscope’s forward kinematics and synchronized with the video streams. This dataset includes 16 videos captured from three pigs (P1, P2, and P3) and three human subjects (H1, H2, and H3), with tissue deformations caused by breathing and manipulation. For our experiments, we select two sequences containing 1,000 and 1,500 consecutive frames, each with a resolution of 512$\times$640. The camera moves around the deformable tissue (see Fig.~\ref{fig:fig1}(b)). We use the provided surgical tool masks to address occlusions. Only left-view images are used for training and evaluation.

\textbf{EndoMapper dataset.} The EndoMapper dataset~\cite{endomapper} is an open-source collection widely used in medical SLAM research, comprising 59 video sequences. In our experiments, we use the colon deformation subset generated with the VR-Caps simulator~\cite{vr-caps}. This subset simulates a forward colonoscope insertion procedure (see Fig.~\ref{fig:fig1}(c)), where colon deformation is modeled using a sine wave: \( V^t_y = V^0_y + A \sin(\omega t + V^0_x + V^0_y + V^0_z) \), where \( V^0_x \), \( V^0_y \), and \( V^0_z \) are the coordinates of surface points at rest. The amplitude \( A \) and frequency \( \omega \) control the extent of deformation. {\color{black}Each sequence also provides ground-truth camera poses that are directly obtained from the simulator.} We use three synthetic sequences, Sequence1, Sequence3, and Sequence5. For convenience, these sequences are referred to as Sequence1, Sequence2, and Sequence3, respectively, in the following experiments. After removing images with camera poses that do not clearly correspond to the visual content, each sequence contains 267 images at a resolution of 512$\times$640.

\subsection{Baseline methods}

To evaluate the performance of our method, we compare it with state-of-the-art approaches for deformable scene reconstruction in endoscopic environments. These include EndoNeRF~\cite{endonerf}, EndoSurf~\cite{endosurf}, Forplane~\cite{yang2024efficient}, Endo-GS~\cite{zhu2024endogs}, Deform3DGS~\cite{yang2024deform3dgs}, SurgicalGaussian~\cite{xie2024surgicalgaussian}, LGS~\cite{liu2024lgs}, EndoGaussian~\cite{liu2025foundation}, EH-SurGS~\cite{eh-surgs}, and DDS-SLAM~\cite{dds-slam}. Among these, EndoNeRF, EndoSurf, and Forplane use implicit neural representations. Endo-GS, Deform3DGS, SurgicalGaussian, LGS, EndoGaussian, and EH-SurGS are based on 3DGS for deformable surgical scene reconstruction. {\color{black}Most of these methods are originally developed for the EndoNeRF and StereoMIS datasets rather than for EndoMapper. To ensure a fair and comprehensive comparison, we additionally include three baselines on the EndoMapper dataset: ENeRF-SLAM~\cite{enerf-slam}, EndoGSLAM~\cite{endogslam}, and Endo-2DTAM~\cite{huang2025advancing}, which represent the current state-of-the-art approaches for colonoscopy video reconstruction. For all SLAM-based methods, including DDS-SLAM, ENeRF-SLAM, EndoGSLAM, and Endo-2DTAM,} we disable the tracking thread and provide the ground-truth camera poses as input, focusing solely on reconstruction performance.  All baseline methods are reproduced using their official repositories with the same RGB and depth inputs.

\begin{table}[t]
\caption{
Depth error and accuracy metrics used for evaluation. Here, $d$ and $d^*$ denote the predicted and ground-truth depth values, respectively, and $D$ represents the set of predicted depth values.
}
  \label{tab:depth_metrics}
  \begin{tabularx}{\columnwidth}{@{}c *{1}{>{\centering\arraybackslash}X}@{}}
  \hline
  Metric & Definition \\
  \hline
  Abs Rel & \( \frac{1}{|D|} \sum_{d \in D} \frac{|d - d^*|}{d^*} \) \\
  Sq Rel & \( \frac{1}{|D|} \sum_{d \in D} \frac{(d - d^*)^2}{d^*} \) \\
  RMSE & \( \sqrt{ \frac{1}{|D|} \sum_{d \in D} (d - d^*)^2 } \) \\
  RMSE log & \( \sqrt{ \frac{1}{|D|} \sum_{d \in D} (\log d - \log d^*)^2 } \) \\
  Accuracy ($\delta < 1.25$) & \( \frac{1}{|D|} \sum_{d \in D} \mathbb{I} \left( \max \left( \frac{d}{d^*},\ \frac{d^*}{d} \right) < 1.25 \right) \) \\
  Accuracy ($\delta < 1.25^2$) & \( \frac{1}{|D|} \sum_{d \in D} \mathbb{I} \left( \max \left( \frac{d}{d^*},\ \frac{d^*}{d} \right) < 1.25^2 \right) \) \\
  \hline
  \end{tabularx}
% \vspace{-5pt}
\end{table}

\begin{table*}[t]
\caption{
Quantitative comparison of different methods on the EndoNeRF dataset. Columns highlighted in \colorbox{blue!10}{\phantom{xx}} indicate that higher values are better, while those in \colorbox{red!10}{\phantom{xx}} indicate that lower values are better. The best results are highlighted in \textbf{bold}, and the second best results are \underline{underlined}.
}
\label{tab:EndoNeRF-all}
\centering
\renewcommand\arraystretch{1.0}
\setlength{\tabcolsep}{3pt}
\resizebox{\textwidth}{!}{
\begin{tabularx}{\textwidth}{l
>{\centering\arraybackslash}X
>{\centering\arraybackslash}X
>{\centering\arraybackslash}X
>{\centering\arraybackslash}X
>{\centering\arraybackslash}X
>{\centering\arraybackslash}X
>{\centering\arraybackslash}X
>{\centering\arraybackslash}X
>{\centering\arraybackslash}X
}
\hline
Method & \cellcolor{blue!10}PSNR & \cellcolor{blue!10}SSIM & \cellcolor{red!10}LPIPS & \cellcolor{red!10}Abs Rel & \cellcolor{red!10}Sq Rel & \cellcolor{red!10}RMSE & \cellcolor{red!10}RMSE log & \cellcolor{blue!10}$\delta < 1.25$ & \cellcolor{blue!10}{$\delta < 1.25^2$} \\
\hline
\multicolumn{10}{l}{\textcolor{gray}{\textit{Pulling}}} \\
EndoNeRF           & 37.723 & 0.949 & 0.088 & 0.354 & 8.500 & 22.607 & 0.390 & 0.421 & 0.746 \\
EndoSurf           & 37.117 & 0.950 & 0.108 & 9.854 & 943.824 & 39.685 & 1.049 & 0.325 & 0.583 \\
ForPlane           & 30.555 & 0.896 & 0.104 & 0.313 & 8.802 & 26.267 & 0.344 & 0.425 & 0.787 \\
DDS-SLAM           & 25.508 & 0.797 & 0.371 & 0.431 & 9.363 & 18.062 & 0.448 & 0.425 & 0.671 \\
EndoGS             & 25.663 & 0.852 & 0.355 & \underline{0.126} & 4.272 & 15.771 & \underline{0.176}   & 0.807 & 0.927 \\
SurgicalGaussian   & 25.805 & 0.853 & 0.368 & 0.177 & 3.557 & 18.457 & 0.185   & 0.837 & 0.908 \\
LGS                & 26.580 & 0.907 & 0.328 & 0.251 & 4.480 & 16.237 & 0.189   & 0.828 & 0.911 \\
EndoGaussian       & 37.190 & 0.955 & 0.066 & {0.171} & \underline{2.568} & \underline{12.012} & 0.195   & 0.816 & \underline{0.936} \\
Deform3DGS         & 37.987 & 0.959 & 0.070 & 0.219 & 2.713 & 12.339 & 0.183   & 0.861 & 0.917 \\
EH-SurGS           & \underline{38.433} & \underline{0.961} & \underline{0.064} & 0.212 & 2.658 & {12.311} & {0.181}   & \underline{0.871} & 0.925 \\
Local-EndoGS       & \textbf{38.727} & \textbf{0.964} & \textbf{0.053} & \textbf{0.119} & \textbf{1.648} & \textbf{6.933} & \textbf{0.147} & \textbf{0.915} & \textbf{0.988} \\
\hline
\multicolumn{10}{l}{\textcolor{gray}{\textit{Cutting}}} \\
EndoNeRF           & 35.962 & 0.936 & 0.094 & 0.323 & 8.836 & 24.244 & 0.348 & 0.410 & 0.803 \\
EndoSurf           & 34.942 & 0.938 & 0.119 & 3.654 & 267.044 & 31.563 & 0.799 & 0.389 & 0.585 \\
ForPlane           & 25.951 & 0.833 & 0.122 & 0.503 & 14.695 & 22.967 & 0.421 & 0.427 & 0.747 \\
DDS-SLAM           & 26.415 & 0.783 & 0.382 & 0.366 & 6.050 & 16.182 & 0.417 & 0.324 & 0.617 \\
EndoGS             & 24.257 & 0.820 & 0.378 & 0.296 & 3.686 & 14.587 & 0.274   & 0.667 & 0.843 \\
SurgicalGaussian   & 25.077 & 0.837 & 0.354 & \underline{0.175} & \underline{2.347} & \underline{10.536} & \underline{0.163}   & \underline{0.823} & \underline{0.942} \\
LGS                & 25.120 & 0.894 & 0.342 & 0.357 & 4.565 & 17.279 & 0.371   & 0.562 & 0.685 \\
EndoGaussian       & 38.040 & 0.960 & 0.052 & 0.332 & 4.121 & 15.815 & 0.291   & 0.642 & 0.778 \\
Deform3DGS         & 37.923 & 0.961 & 0.054 & 0.236 & 2.812 & 12.475 & 0.200   & 0.729 & 0.894 \\
EH-SurGS           & \underline{39.457} & \underline{0.967} & \underline{0.039} & 0.233 & 2.748 & 12.107 & 0.196   & 0.738 & 0.908 \\
Local-EndoGS       & \textbf{39.647} & \textbf{0.968} & \textbf{0.037} & \textbf{0.107} & \textbf{1.136} & \textbf{5.825} & \textbf{0.135} & \textbf{0.905} & \textbf{0.973} \\
\hline
\end{tabularx}
}
% \vspace{-5pt}
\end{table*}
% \FloatBarrier

\subsection{Evaluation metrics}

Following previous work~\cite{endosurf,eh-surgs,uc-nerf}, we report quantitative comparisons for both appearance and geometry. For appearance quality, we use standard metrics: Peak Signal-to-Noise Ratio (PSNR), Structural Similarity Index (SSIM)~\cite{ssim}, and Learned Perceptual Image Patch Similarity (LPIPS)~\cite{lpips}. For geometry, we assess depth prediction accuracy using common metrics from prior studies~\cite{shao2022self,uc-nerf}: Absolute Relative Error (Abs Rel), Squared Relative Error (Sq Rel), Root Mean Square Error (RMSE), RMSE log, and accuracy under the threshold \( \delta < t \), where \( t \in \{1.25, 1.25^2\} \). The definitions of all error and accuracy metrics are summarized in Table~\ref{tab:depth_metrics}. Following standard procedure for monocular depth estimation, we apply median scaling during evaluation. The scaling factor is computed as the ratio of the medians of the ground-truth and predicted depth maps.

\subsection{Implementation Details}

We implement our method in PyTorch. All experiments are conducted on an Ubuntu 20.04 system with a single RTX 4090 GPU and an Intel Xeon Platinum 8474 CPU. We use the Adam optimizer with an initial learning rate of \( 1.6 \times 10^{-3} \), and other training parameters are set as in the original 3DGS. Each local window is trained for 1,000 iterations. Following previous work~\cite{endosurf,shan2025uw}, we split each dataset into training and testing sets using a 7:1 ratio. {\color{black}We quantify translational difference using the {mean squared difference (MSE)} between translation vectors. 
For the StereoMIS, the translational threshold is set to $\delta_t = 0.6\,\text{cm}^2$, corresponding to an approximate physical displacement of $7.7$\,cm. 
For the EndoMapper, $\delta_t = 0.6\,\text{mm}^2$, corresponding to about $7.7$\,mm of displacement. 
For both datasets, the rotational threshold is fixed at $\delta_r = 15^\circ$.} The number of windows for each sequence, determined by our adaptive window partitioning method, is summarized in Table~\ref{tab:dataset_summary}. To ensure fairness, we run all methods on each dataset three times and report the average results.

\begin{table*}[t]
\caption{
Quantitative comparison of different methods on the StereoMIS dataset. Columns highlighted in \colorbox{blue!10}{\phantom{xx}} indicate that higher values are better, while those highlighted in \colorbox{red!10}{\phantom{xx}} indicate that lower values are better. The best results are shown in \textbf{bold}, and the second-best results are \underline{underlined}.
}
\label{tab:StereoMIS-all}
\centering
\renewcommand\arraystretch{1.0}
\setlength{\tabcolsep}{2pt}
\resizebox{\textwidth}{!}{
\begin{tabularx}{\textwidth}{l
>{\centering\arraybackslash}X
>{\centering\arraybackslash}X
>{\centering\arraybackslash}X
>{\centering\arraybackslash}X
>{\centering\arraybackslash}X
>{\centering\arraybackslash}X
>{\centering\arraybackslash}X
>{\centering\arraybackslash}X
>{\centering\arraybackslash}X
}
\hline
Method & \cellcolor{blue!10}PSNR & \cellcolor{blue!10}SSIM & \cellcolor{red!10}LPIPS & \cellcolor{red!10}Abs Rel & \cellcolor{red!10}Sq Rel & \cellcolor{red!10}RMSE & \cellcolor{red!10}RMSE log & \cellcolor{blue!10}$\delta < 1.25$ & \cellcolor{blue!10}{$\delta < 1.25^2$} \\
\hline
\multicolumn{10}{l}{\textcolor{gray}{\textit{StereoMIS-Sequence1}}} \\
EndoNeRF & \underline{25.913} & \underline{0.746} & \underline{0.481} & 0.420 & 23.226 & 58.555 & 0.478 & 0.310 & 0.590 \\
EndoSurf & 24.883 & 0.715 & 0.520 & 3.219 & 472.041 & 51.691 & 0.736 & 0.319 & 0.578 \\
ForPlane & 16.389 & 0.651 & 0.792 & 0.640 & 36.502 & 62.023 & 0.613 & 0.299 & 0.544 \\
DDS-SLAM & 16.585 & 0.569 & 0.673 & \underline{0.355} & 20.031 & 40.680 & \underline{0.402} & 0.462 & 0.726 \\
EndoGS & 20.412 & 0.691 & 0.643 & 0.572 & \underline{10.705} & 29.252 & 0.522 & 0.454 & \underline{0.740} \\
SurgicalGaussian & 21.149 & 0.714 & 0.558 & 0.678 & 12.574 & 32.563 & 0.596 & 0.396 & 0.611 \\
LGS & 18.540 & 0.560 & 0.620 & 0.753 & 15.270 & 34.863 & 0.718 & 0.326 & 0.524 \\
EndoGaussian & 20.440 & 0.690 & 0.650 & 0.618 & 11.384 & 31.157 & 0.553 & 0.453 & 0.669 \\
Deform3DGS & 20.497 & 0.697 & 0.646 & 0.582 & 10.928 & 27.536 & 0.541 & 0.513 & 0.707 \\
EH-SurGS & 21.161 & 0.716 & 0.556 & 0.574 & 10.754 & \underline{27.419} & 0.529 & \underline{0.531} & 0.724 \\
Local-EndoGS & \textbf{32.294} & \textbf{0.892} & \textbf{0.304} & \textbf{0.112} & \textbf{2.951} & \textbf{7.485} & \textbf{0.149} & \textbf{0.919} & \textbf{0.976} \\
\hline
\multicolumn{10}{l}{\textcolor{gray}{\textit{StereoMIS-Sequence2}}} \\
EndoNeRF & 11.857 & 0.433 & 0.773 & 1.632 & 123.563 & 81.365 & 0.901 & 0.157 & 0.313 \\
EndoSurf & 23.614 & 0.601 & 0.677 & {0.831} & 52.405 & 33.187 & 0.593 & 0.359 & 0.622 \\
ForPlane & 23.123 & 0.589 & 0.586 & 1.978 & 86.499 & 60.861 & 0.819 & 0.209 & 0.436 \\
DDS-SLAM & 19.577 & 0.523 & 0.662 & \underline{0.280} & \underline{11.284} & \underline{29.105} & \underline{0.333} & \underline{0.557} & \underline{0.828} \\
EndoGS & 15.493 & 0.635 & 0.692 & 1.254 & 55.003 & 42.173 & 0.797 & 0.277 & 0.504 \\
SurgicalGaussian & 20.940 & 0.717 & 0.603 & 1.807 & 59.938 & 45.967 & 0.869 & 0.197 & 0.392 \\
LGS & 22.393 & 0.733 & 0.586 & 2.213 & 78.488 & 57.582 & 1.130 & 0.105 & 0.218 \\
EndoGaussian & \underline{23.927} & \underline{0.753} & \underline{0.575} & 1.774 & 58.630 & 42.205 & 0.826 & 0.242 & 0.471 \\
Deform3DGS & 21.453 & 0.719 & 0.598 & 1.217 & 43.830 & 35.641 & 0.728 & 0.293 & 0.551 \\
EH-SurGS & 22.560 & 0.741 & 0.587 & 1.201 & {43.173} & 35.174 & 0.719 & 0.298 & 0.569 \\
Local-EndoGS & \textbf{31.487} & \textbf{0.922} & \textbf{0.297} & \textbf{0.129} & \textbf{3.852} & \textbf{9.271} & \textbf{0.154} & \textbf{0.912} & \textbf{0.967} \\
\hline
\end{tabularx}
}
\end{table*}

% \afterpage{%
\begin{figure*}[!htbp]
\centerline{\includegraphics[width=0.99\linewidth]{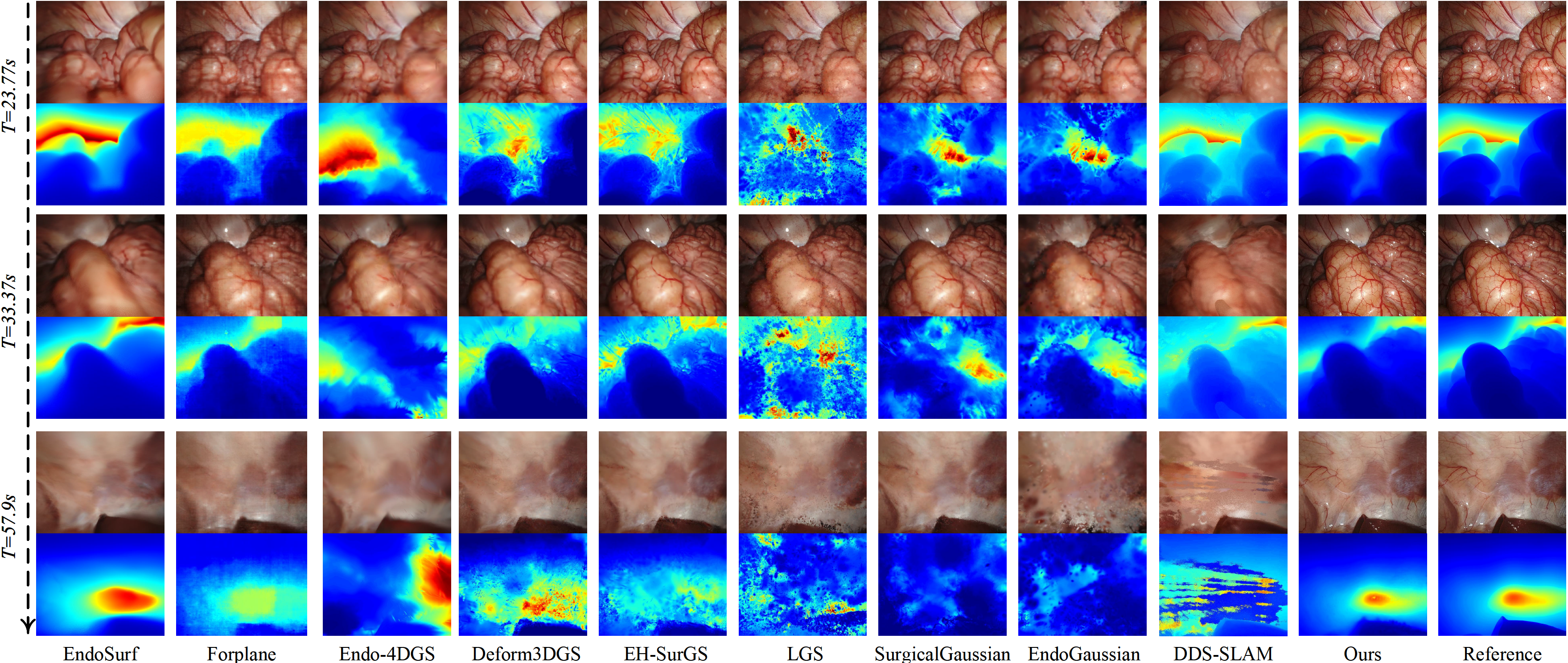}}
\caption{
Qualitative comparison of image rendering and depth prediction on deformable scenes at different time points from the StereoMIS dataset. Each pair of rows represents a specific time point: the first row shows the rendered RGB images, and the second row shows the predicted depth maps. The figure presents results from three time points (from top to bottom), illustrating how the observed scene changes as the camera moves. Compared to existing methods, Local-EndoGS (Ours) consistently provides finer reconstruction details, while the baseline methods show limitations in both image quality and depth accuracy.
}
\label{fig:stereomis}
\end{figure*}
% }

\begin{table*}[!t]
\centering
\caption{
Quantitative comparison of various methods on the EndoMapper dataset.
Columns highlighted in \colorbox{blue!10}{\phantom{xx}} indicate that higher values are better,
while those highlighted in \colorbox{red!10}{\phantom{xx}} indicate that lower values are better.
The best results are shown in \textbf{bold}, and the second-best results are \underline{underlined}.
}
\label{tab:EndoMapper-all}

\renewcommand\arraystretch{1.0}
\setlength{\tabcolsep}{2pt}

\resizebox{\textwidth}{!}{%
\begin{tabularx}{\textwidth}{
l
>{\centering\arraybackslash}X  % PSNR
>{\centering\arraybackslash}X  % SSIM
>{\centering\arraybackslash}X  % LPIPS
>{\centering\arraybackslash}X  % Abs Rel
>{\centering\arraybackslash}X  % Sq Rel
>{\centering\arraybackslash}X  % RMSE
>{\centering\arraybackslash}X  % RMSE log
>{\centering\arraybackslash}X  % delta<1.25
>{\centering\arraybackslash}X  % delta<1.25^2
}
\hline
Method & \cellcolor{blue!10}PSNR & \cellcolor{blue!10}SSIM & \cellcolor{red!10}LPIPS & \cellcolor{red!10}Abs Rel & \cellcolor{red!10}Sq Rel & \cellcolor{red!10}RMSE & \cellcolor{red!10}RMSE log & \cellcolor{blue!10}$\delta < 1.25$ & \cellcolor{blue!10}{$\delta < 1.25^2$} \\
\hline
\multicolumn{10}{l}{\textcolor{gray}{\textit{EndoMapper-Sequence1}}} \\
EndoNeRF & 11.192 & 0.626 & 0.824 & \underline{0.401} & 30.472 & 64.341 & 0.602 & 0.301 & 0.589 \\
EndoSurf & 23.767 & 0.689 & 0.714 & 7.182 & 740.954 & 62.653 & 0.962 & 0.258 & 0.494 \\
ForPlane & \underline{27.026} & \underline{0.811} & 0.552 & 0.825 & 24.607 & 49.890 & 0.622 & 0.283 & 0.603 \\
DDS-SLAM & 22.315 & 0.739 & 0.643 & 0.443 & 28.178 & 40.461 & 0.594 & 0.364 & 0.623 \\
\textcolor{black}{ENeRF-SLAM}
& \textcolor{black}{19.661} & \textcolor{black}{0.757} & \textcolor{black}{0.609} & \textcolor{black}{0.499} & \textcolor{black}{31.652} & \textcolor{black}{44.688} & \textcolor{black}{0.700} & \textcolor{black}{0.298} & \textcolor{black}{0.582} \\
{\color{black}EndoGSLAM} & \textcolor{black}{17.452} & \textcolor{black}{0.675} & \textcolor{black}{0.551} & \textcolor{black}{0.971} & \textcolor{black}{88.893} & \textcolor{black}{63.412} & \textcolor{black}{0.886} & \textcolor{black}{0.197} & \textcolor{black}{0.392} \\
{\color{black}Endo-2DTAM} & \textcolor{black}{15.509} & \textcolor{black}{0.583} & \textcolor{black}{0.622} & \textcolor{black}{0.988} & \textcolor{black}{90.022} & \textcolor{black}{62.122} & \textcolor{black}{0.869} & \textcolor{black}{0.209} & \textcolor{black}{0.413} \\
Endo-4DGS & 19.453 & 0.714 & 0.682 & 0.615 & 9.388 & 28.909 & 0.630 & 0.786 & 0.930 \\
SurgicalGaussian & 20.450 & 0.720 & 0.572 & 0.669 & 10.803 & 29.715 & 0.680 & 0.738 & 0.872 \\
LGS & 23.560 & 0.772 & 0.555 & 0.713 & 12.597 & 30.475 & 0.692 & 0.655 & 0.848 \\
EndoGaussian & 25.137 & 0.792 & 0.583 & 0.583 & \underline{8.911} & \underline{25.999} & \underline{0.441} & 0.759 & 0.913 \\
Deform3DGS & 24.767 & 0.788 & \underline{0.539} & 0.615 & 9.388 & 28.909 & 0.630 & 0.786 & 0.930 \\
EH-SurGS & 25.477 & 0.803 & 0.572 & 0.610 & 9.373 & 28.722 & 0.618 & \underline{0.794} & \underline{0.936} \\
Local-EndoGS & \textbf{33.483} & \textbf{0.944} & \textbf{0.198} & \textbf{0.136} & \textbf{3.265} & \textbf{8.422} & \textbf{0.152} & \textbf{0.903} & \textbf{0.973} \\
\hline
\multicolumn{10}{l}{\textcolor{gray}{\textit{EndoMapper-Sequence2}}} \\
EndoNeRF & 11.032 & 0.625 & 0.813 & 0.784 & 72.588 & 65.078 & 0.857 & 0.195 & 0.393 \\
EndoSurf & 24.445 & 0.703 & 0.703 & 8.066 & 661.562 & 59.846 & 1.013 & 0.267 & 0.510 \\
ForPlane & \underline{27.075} & \underline{0.811} & 0.554 & 1.096 & 28.376 & 47.761 & 0.626 & 0.297 & 0.622 \\
DDS-SLAM & 21.916 & 0.747 & 0.631 & \underline{0.444} & 27.051 & 39.969 & 0.626 & 0.358 & 0.614 \\
\textcolor{black}{ENeRF-SLAM} & \textcolor{black}{19.118} & \textcolor{black}{0.755} & \textcolor{black}{0.611} & \textcolor{black}{0.512} & \textcolor{black}{31.669} & \textcolor{black}{44.348} & \textcolor{black}{0.731} & \textcolor{black}{0.284} & \textcolor{black}{0.560} \\
\textcolor{black}{EndoGSLAM} & \textcolor{black}{17.290} & \textcolor{black}{0.677} & \textcolor{black}{0.562} & \textcolor{black}{0.903} & \textcolor{black}{83.272} & \textcolor{black}{64.273} & \textcolor{black}{0.882} & \textcolor{black}{0.203} & \textcolor{black}{0.387} \\
\textcolor{black}{Endo-2DTAM} & \textcolor{black}{15.686} & \textcolor{black}{0.592} & \textcolor{black}{0.615} & \textcolor{black}{0.962} & \textcolor{black}{83.923} & \textcolor{black}{62.772} & \textcolor{black}{0.868} & \textcolor{black}{0.209} & \textcolor{black}{0.399} \\
Endo-4DGS & 17.827 & 0.673 & 0.697 & 0.635 & 10.086 & 31.252 & 0.677 & 0.742 & 0.847 \\
SurgicalGaussian & 20.033 & 0.708 & 0.628 & 0.696 & 11.148 & 33.063 & 0.684 & 0.725 & 0.853 \\
LGS & 23.053 & 0.752 & 0.579 & 0.728 & 12.809 & 35.726 & 0.708 & 0.696 & 0.833 \\
EndoGaussian & 24.870 & 0.790 & 0.586 & 0.595 & \underline{9.317} & \underline{29.562} & \underline{0.459} & \underline{0.767} & \underline{0.892} \\
Deform3DGS & 24.250 & 0.773 & \underline{0.511} & 0.635 & 10.086 & 31.252 & 0.677 & 0.742 & 0.847 \\
EH-SurGS & 24.847 & 0.795 & 0.566 & 0.628 & 10.055 & 31.109 & 0.661 & 0.747 & 0.864 \\
Local-EndoGS & \textbf{32.993} & \textbf{0.940} & \textbf{0.217} & \textbf{0.145} & \textbf{3.640} & \textbf{9.542} & \textbf{0.163} & \textbf{0.885} & \textbf{0.961} \\
\hline
\multicolumn{10}{l}{\textcolor{gray}{\textit{EndoMapper-Sequence3}}} \\
EndoNeRF & 9.703 & 0.572 & 0.825 & 0.898 & 88.927 & 67.545 & 0.912 & 0.186 & 0.365 \\
EndoSurf & 25.618 & 0.748 & 0.663 & 7.947 & 1013.194 & 56.154 & 0.906 & 0.262 & 0.499 \\
ForPlane & \underline{26.979} & \underline{0.814} & 0.569 & 1.077 & 24.212 & 44.934 & 0.605 & 0.335 & 0.646 \\
DDS-SLAM & 22.315 & 0.739 & 0.643 & \underline{0.469} & 31.711 & 41.428 & 0.624 & 0.347 & 0.603 \\
\textcolor{black}{ENeRF-SLAM} & \textcolor{black}{18.869} & \textcolor{black}{0.759} & \textcolor{black}{0.619} & \textcolor{black}{0.541} & \textcolor{black}{32.177} & \textcolor{black}{44.780} & \textcolor{black}{0.751} & \textcolor{black}{0.249} & \textcolor{black}{0.519} \\
\textcolor{black}{EndoGSLAM} & \textcolor{black}{11.755} & \textcolor{black}{0.665} & \textcolor{black}{0.566} & \textcolor{black}{1.000} & \textcolor{black}{93.154} & \textcolor{black}{63.095} & \textcolor{black}{0.901} & \textcolor{black}{0.204} & \textcolor{black}{0.392} \\
\textcolor{black}{Endo-2DTAM} & \textcolor{black}{15.120} & \textcolor{black}{0.579} & \textcolor{black}{0.621} & \textcolor{black}{0.998} & \textcolor{black}{87.916} & \textcolor{black}{61.317} & \textcolor{black}{0.876} & \textcolor{black}{0.214} & \textcolor{black}{0.411} \\
Endo-4DGS & 19.453 & 0.714 & 0.682 & 0.619 & 9.671 & 29.755 & 0.658 & 0.764 & 0.918 \\
SurgicalGaussian & 20.450 & 0.720 & 0.572 & 0.652 & 9.972 & 30.933 & 0.675 & 0.761 & 0.896 \\
LGS & 23.560 & 0.772 & 0.555 & 0.659 & 11.288 & 33.825 & 0.687 & 0.679 & 0.842 \\
EndoGaussian & 25.137 & 0.792 & 0.583 & {0.601} & \underline{9.344} & \underline{27.165} & \underline{0.431} & \underline{0.783} & \underline{0.961} \\
Deform3DGS & 24.767 & 0.788 & \underline{0.539} & 0.619 & 9.671 & 29.755 & 0.658 & 0.764 & 0.918 \\
EH-SurGS & 25.477 & 0.803 & 0.572 & 0.612 & 9.652 & 29.602 & 0.639 & 0.774 & 0.932 \\
Local-EndoGS & \textbf{33.483} & \textbf{0.944} & \textbf{0.198} & \textbf{0.141} & \textbf{3.360} & \textbf{8.969} & \textbf{0.159} & \textbf{0.898} & \textbf{0.970} \\
\hline
\end{tabularx}}
\end{table*}

\section{Experimental results}

\begin{figure*}[t]
\centerline{\includegraphics[width=0.99\linewidth]{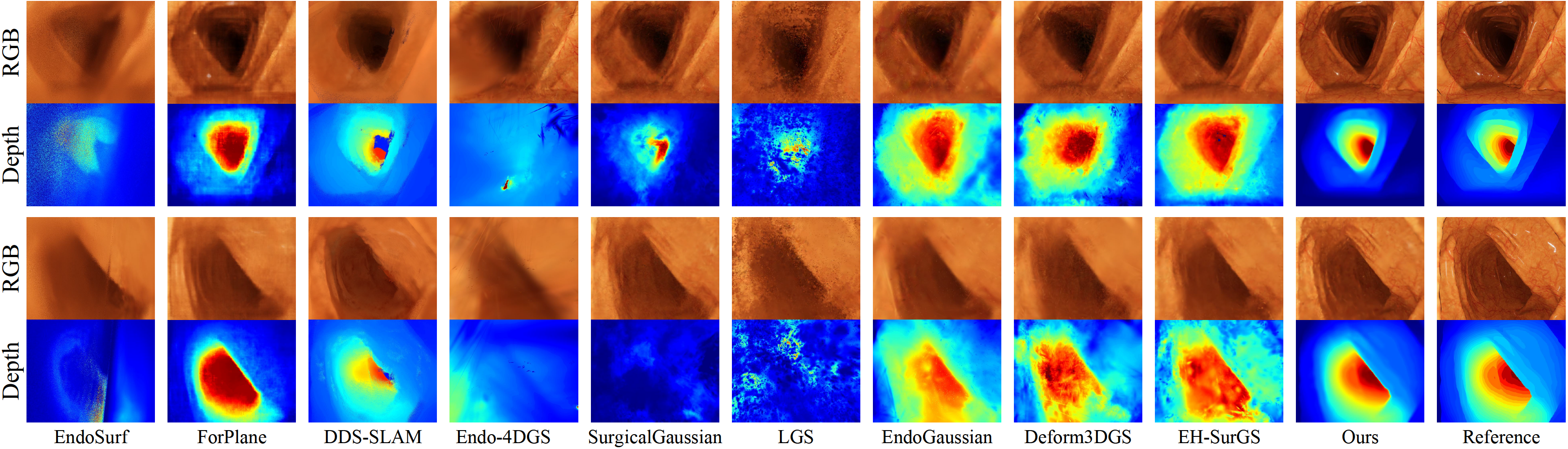}}
    % \vspace{-4pt}
\caption{Qualitative comparison of appearance (RGB) and geometric (depth) reconstruction results produced by Local-EndoGS and existing methods on deformable scenes from the EndoMapper dataset. Our method produces more accurate and visually consistent results for both appearance and depth, with fewer artifacts and better delineation of anatomical structures compared to existing approaches.}
    % \vspace{-15pt}
    \label{fig:endomapper}
\end{figure*}

Quantitative results on the three datasets are summarized in Table~\ref{tab:EndoNeRF-all}, Table~\ref{tab:StereoMIS-all}, and Table~\ref{tab:EndoMapper-all}. Our method consistently achieves higher appearance rendering quality and more accurate depth prediction than existing approaches across all datasets. Qualitative results, shown in Fig.~\ref{fig:stereomis} and
Fig.~\ref{fig:endomapper}, further confirm these findings by demonstrating that our approach produces visually realistic renderings and accurate depth maps under various camera motions. Overall, these results demonstrate the effectiveness and robustness of our method for 4D monocular surgical reconstruction.

\subsection{Results on the EndoNeRF Dataset}

As shown in Table~\ref{tab:EndoNeRF-all}, on the EndoNeRF dataset with a fixed camera view, Local-EndoGS achieved the best performance on both sequences. In the \textit{Pulling} and \textit{Cutting} sequences, Local-EndoGS obtained the highest scores for image quality metrics such as PSNR and SSIM (38.727/0.964 and 39.647/0.968, respectively), and the lowest LPIPS values (0.053 and 0.037), indicating better image reconstruction quality. For geometric reconstruction, baseline methods rely heavily on stereo depth as an additional input. When only monocular sequences are available, the scale ambiguity of monocular depth priors leads to significant errors in geometric reconstruction. Our method addresses this problem and substantially outperforms baseline methods in depth prediction. For example, on the \textit{Pulling} sequence, Local-EndoGS improved Abs Rel, Sq Rel, RMSE, and RMSE log by 5.6\%, 35.8\%, 42.3\%, and 16.5\%, respectively, compared with the second-best method. For accuracy metrics ($\delta<1.25$ and $\delta<1.25^2$), the improvements were 5.0\% and 5.6\%. The improvements on the \textit{Cutting} sequence were even more significant, with error metric improvements ranging from 17.2\% to 51.6\%, and accuracy metrics improved by 10.0\% and 3.3\%. These results demonstrate that our approach effectively preserves high-fidelity appearance while significantly improving geometric accuracy compared to existing methods.

\begin{table*}[t]
\caption{Training Time and Rendering Speed on Different Datasets}
\label{tab:efficiency_time_speed}
\centering
\begin{tabularx}{0.88\textwidth}{>{\hspace{0.5em}}l|>{\centering\arraybackslash}X>{\centering\arraybackslash}X!{\vrule}
>{\centering\arraybackslash}X>{\centering\arraybackslash}X}
\toprule
\multirow{2}{*}{Method} 
    & \multicolumn{2}{c!{\vrule}}{\textbf{EndoNeRF dataset}} 
    & \multicolumn{2}{c}{\textbf{StereoMIS dataset}} \\
    & Train Time & Rendering Speed & Train Time & Rendering Speed \\
\midrule
EndoNeRF         & $\sim$ 8 h   & 0.18 fps      & $\sim$ 18 h  & 0.13 fps \\
EndoSurf         & $\sim$ 8 h   & 0.18 fps      & $\sim$ 10 h  & 0.26 fps \\
Forplane         & 3.00 min     & 1.40 fps      & 2.80 min     & 1.30 fps \\
Endo-GS          & 4.43 min     & 118.50 fps    & 6.55 min     & 85.33 fps \\
Deform3DGS       & 1.39 min     & 354.80 fps    & 3.28 min     & 202.17 fps \\
SurgicalGaussian & 2.74 min     & 159.33 fps    & 4.49 min     & 114.00 fps \\
LGS              & 1.97 min     & 135.50 fps    & 3.70 min     & 86.83 fps \\
EndoGaussian     & 2.89 min     & 203.67 fps    & 4.14 min     & 221.50 fps \\
EH-SurGS         & 1.66 min     & 371.33 fps    & 3.67 min     & 215.33 fps \\
Local-EndoGS     & 2.38 min     & 371.00 fps    & 8.36 min     & 329.83 fps \\
\bottomrule
\end{tabularx}
\end{table*}
\subsection{Results on the StereoMIS Dataset}

Unlike the EndoNeRF dataset, where the camera view is fixed, the StereoMIS dataset involves a moving camera around the tissue. This movement introduces additional challenges for deformable reconstruction. As shown in Table~\ref{tab:StereoMIS-all}, all baseline methods show a clear decrease in appearance rendering quality. For example, the PSNR of EndoGaussian~\cite{liu2025foundation} drops from 37.190~dB on the EndoNeRF-\textit{Pulling} sequence (Table~\ref{tab:EndoNeRF-all}) to 20.440~dB on StereoMIS-\textit{Sequence1}. This decrease is mainly due to camera movement, which breaks the static camera assumption and causes inconsistencies between the observed and canonical spaces. Local-EndoGS consistently achieves the highest PSNR and SSIM, with improvements over the second-best method of 24.1\% and 17.1\% in \textit{Sequence1}, and 31.6\% and 22.4\% in \textit{Sequence2}. For perceptual similarity (LPIPS), Local-EndoGS reduces the error by 36.8\% in \textit{Sequence1} and 48.3\% in \textit{Sequence2} compared to the next best method, indicating an advantage in both image synthesis fidelity and perceptual quality. For geometric reconstruction, Local-EndoGS achieves substantial improvements across all error metrics (Abs Rel, Sq Rel, RMSE, and RMSE log). In \textit{Sequence1}, improvements are 68.5\%, 72.4\%, 72.7\%, and 62.9\%, respectively. In \textit{Sequence2}, the improvements are 53.9\%, 65.8\%, 68.2\%, and 53.8\%. For accuracy metrics ($\delta < 1.25$ and $\delta < 1.25^2$), \textit{Sequence1} shows improvements of 73.0\% and 31.9\%, and \textit{Sequence2} shows improvements of 63.8\% and 16.8\%.

We also provide a qualitative comparison of different methods on the StereoMIS dataset at three time points, illustrating how the observed deformable scene changes as the camera moves, as shown in Fig.~\ref{fig:stereomis}. For appearance rendering, most baseline methods produce visible artifacts or blurring and fail to capture fine tissue textures. In terms of geometric reconstruction, as seen in the depth maps, many baseline methods do not produce reliable results when the camera moves, often resulting in noisy or distorted depth estimates. While implicit neural representation methods (such as EndoSurf, Forplane, and DDS-SLAM) offer some improvements in fitting, they still lack geometric detail. Moreover, these methods require longer training times and slower inference speeds (see Sec.~\ref{sec:efficiency}). In contrast, our Local-EndoGS method preserves finer texture details and generates more accurate geometric structures, enabling faithful reconstruction of both appearance and geometry in deformable surgical environments.

\begin{table*}[t]
\caption{
Ablation Studies of Different Model Components on the StereoMIS dataset.
}
\label{tab:ablation}
\centering
\renewcommand\arraystretch{1.1}
\setlength{\tabcolsep}{3pt}
\resizebox{\textwidth}{!}{
\begin{tabularx}{\textwidth}{l
>{\centering\arraybackslash}X % PSNR
>{\centering\arraybackslash}X % SSIM
>{\centering\arraybackslash}X % Abs Rel
>{\centering\arraybackslash}X % Sq Rel
>{\centering\arraybackslash}X % RMSE
>{\centering\arraybackslash}X % RMSE log
>{\centering\arraybackslash}X % delta<1.25
>{\centering\arraybackslash}X % delta<1.25^2
>{\centering\arraybackslash}X % FPS
}
\hline
Method & \cellcolor{blue!10}PSNR & \cellcolor{blue!10}SSIM & \cellcolor{red!10}Abs Rel & \cellcolor{red!10}Sq Rel & \cellcolor{red!10}RMSE & \cellcolor{red!10}RMSE log & \cellcolor{blue!10}$\delta < 1.25$ & \cellcolor{blue!10}{$\delta < 1.25^2$} & \cellcolor{blue!10}FPS \\
\hline
\multicolumn{10}{l}{\textcolor{gray}{\textit{Progressive Window-based Global Scene Representation (PWGSR)}}} \\
w/o windows & 21.76 & 0.713 & 1.105 & 38.888 & 32.925 & 0.692 & 0.354 & 0.617 & 306 \\
w/o AWP     & 31.09 & 0.910 & 0.137 & 4.621  & 9.674  & 0.170 & 0.891 & 0.949 & 325 \\
\hline
\multicolumn{10}{l}{\textcolor{gray}{\textit{Local Canonical Space Initialization (LCSI)}}} \\
w SD        & 31.77 & 0.923 & 0.124 & 3.765  & 9.249  & 0.153 & 0.917 & 0.975 & 325 \\
w/o LCSI    & 28.38 & 0.885 & 0.296 & 9.858  & 18.794 & 0.426 & 0.657 & 0.814 & 325 \\
w/o TAP     & 27.47 & 0.855 & 0.281 & 9.384  & 14.712 & 0.384 & 0.696 & 0.869 & 330 \\
w/o CWIP    & 31.22 & 0.919 & 0.133 & 3.943  & 9.305  & 0.157 & 0.907 & 0.962 & 325 \\
w/o EGRR    & 30.90 & 0.901 & 0.139 & 4.787  & 9.794  & 0.170 & 0.883 & 0.948 & 329 \\
\hline
\multicolumn{10}{l}{\textcolor{gray}{\textit{Loss Functions (LF)}}} \\
w/o TL      & 31.16 & 0.918 & 0.129 & 3.921  & 9.285  & 0.156 & 0.908 & 0.964 & 326 \\
w/o PBR     & 30.87 & 0.897 & 0.138 & 4.764  & 9.983  & 0.174 & 0.879 & 0.942 & 322 \\
\hline
Full model  & 31.49 & 0.922 & 0.129 & 3.852  & 9.271  & 0.154 & 0.912 & 0.967 & 330 \\
\hline
\end{tabularx}
}
\end{table*}

\subsection{Results on EndoMapper Dataset}

Table~\ref{tab:EndoMapper-all} presents the quantitative results for all methods on the three sequences of the EndoMapper dataset. Consistent with the findings on the StereoMIS dataset, our proposed Local-EndoGS achieves the best performance across all evaluation metrics. For image quality metrics such as PSNR, SSIM, and LPIPS, Local-EndoGS consistently outperforms the second-best method, producing images with fewer artifacts and better preservation of fine details. In terms of depth estimation accuracy—including Abs Rel, RMSE, and RMSE log—Local-EndoGS demonstrates clear improvements, delivering more reliable and accurate geometric reconstructions compared to other approaches. {\color{black}For methods specifically designed for colonoscopy videos, such as EndoGSLAM \cite{endogslam} and Endo-2DTAM \cite{huang2025advancing}, their performance is noticeably inferior since they ignore the deformable nature of the scene.} Qualitative comparisons are shown in Fig.~\ref{fig:endomapper}.

\subsection{Analysis of Model Efficiency}\label{sec:efficiency}

To evaluate model efficiency, we report the training time and inference speed (frames per second, FPS) for each method on the EndoNeRF and StereoMIS datasets in Table~\ref{tab:efficiency_time_speed}. On the EndoNeRF dataset, our method completes training in about two minutes, which is comparable to most 3DGS-based methods, and is suitable for clinical analysis and offline processing. It achieves a real-time rendering speed of 371 FPS, supporting efficient visualization and post-processing. On the StereoMIS dataset, the training time increases due to larger scene sizes (over 2000 frames), wider camera movement, and more complex tissue deformation. These factors require more local windows (as shown in Table~\ref{tab:dataset_summary}) and additional optimization steps. This increase is expected, as more extensive and dynamic surgical scenes demand extra computation to capture local variations and ensure reliable reconstruction. Notably, our method achieves the fastest inference speed, demonstrating high efficiency and scalability.

\begin{figure}[t]
    \centering
\includegraphics[width=0.45\textwidth]{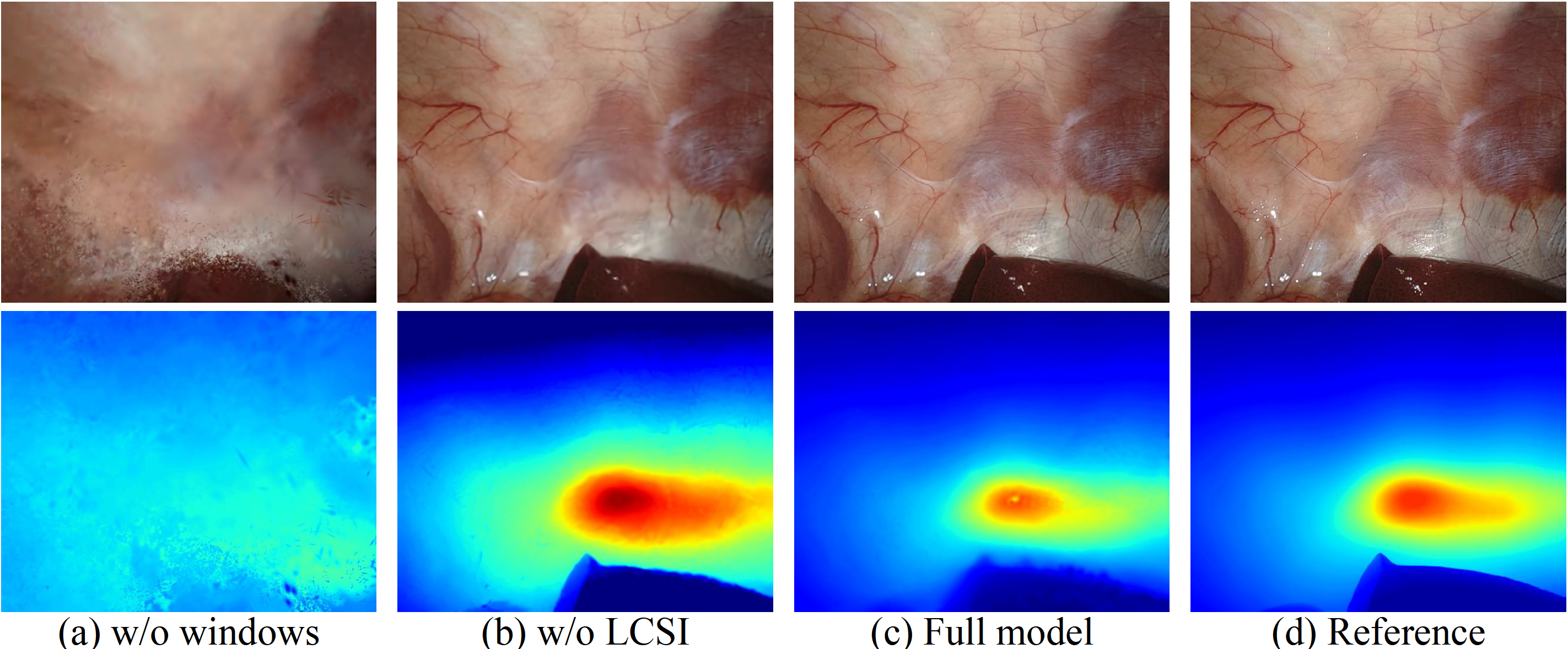}
\caption{
Qualitative comparison of ablation results for RGB reconstruction (top row) and depth estimation (bottom row). (a) Without local windows; (b) Without local canonical space initialization (LCSI); (c) Full model; (d) Reference.
}
\label{fig:ablation_visualization}
% \vspace{-10pt}
\end{figure}

\subsection{Ablation Study}\label{sec:ablation}

In this subsection, we evaluate the effectiveness of each module in our proposed method through comprehensive ablation experiments on StereoMIS-Sequence2. The quantitative results are summarized in Table~\ref{tab:ablation}.

\textbf{Progressive Window-based Global Scene Representation (PWGSR).}
To model long endoscopic sequences with arbitrary camera motion, we use adaptive window partitioning (AWP) to divide the input sequence into multiple local windows. We then progressively optimize the parameters of each local model to represent the scene within each window. To assess the effectiveness of this approach, we conduct three experiments: (1) the \underline{full model}; (2) a variant without window partitioning (\underline{w/o windows}), which uses a single window for the entire scene, similar to previous methods; and (3) a variant that uses uniform window partitioning instead of AWP (\underline{w/o AWP}). The number of windows is kept the same as in the full model to ensure a fair comparison. As shown in Table~\ref{tab:ablation}, removing window partitioning leads to a clear drop in performance because a single canonical space and deformation network cannot capture the full range of scene deformations. Fig.~\ref{fig:ablation_visualization} further demonstrates this effect. The results from the w/o windows variant (Fig.~\ref{fig:ablation_visualization}(a)) show pronounced artifacts and loss of structural detail, while the full model (Fig.~\ref{fig:ablation_visualization}(c)) produces reconstructions that closely match the reference. Similarly, removing the AWP component (w/o AWP) also reduces performance, as uniform partitioning uses a fixed window size and cannot adapt to regions with varying dynamics.

\textbf{Local Canonical Space Initialization (LCSI).}
We conduct six ablation experiments to evaluate the effectiveness of LCSI: (1) the \underline{full model}; (2) initialization using a stereo depth prior, as in existing methods (\underline{w SD}); (3) removing local canonical space initialization and using only a monocular depth prior (\underline{w/o LCSI}); (4) replacing the TAP model with SIFT keypoints \cite{lowe2004distinctive} and brute-force matching from OpenCV to establish inter-frame correspondences (\underline{w/o TAP}); (5) removing cross-window information propagation (\underline{w/o CWIP}); and (6) removing Error-Guided Region Refinement (\underline{w/o EGRR}). As shown in Table~\ref{tab:ablation}, although our full model uses monocular input, it achieves performance comparable to the model using stereo depth priors ({w SD}). This demonstrates the effectiveness of our initialization scheme for monocular sequences. Furthermore, initializing with only a monocular depth prior (\underline{w/o LCSI}) leads to a significant drop in performance, mainly due to scale ambiguity. As shown in Fig.~\ref{fig:ablation_visualization}(b), the reconstructed RGB image has noticeable artifacts, with many anatomical details blurred. In addition, the estimated depth map is less accurate compared to the full model (Fig.~\ref{fig:ablation_visualization}(c)) and the reference (Fig.~\ref{fig:ablation_visualization}(d)). In contrast, our full model produces both RGB and depth results that closely resemble the reference, highlighting the effectiveness of the proposed local canonical space initialization. The results of removing the TAP model show that traditional correspondence methods cannot effectively address the challenges in endoscopic images, underscoring the necessity of the TAP module. Removing CWIP also reduces performance, as it helps maintain spatial consistency between different local windows. Finally, removing EGRR lowers performance, confirming that error-guided region refinement corrects regional inaccuracies and improves overall reconstruction quality.

\begin{figure}[t]
\centerline{\includegraphics[width=0.98\linewidth]{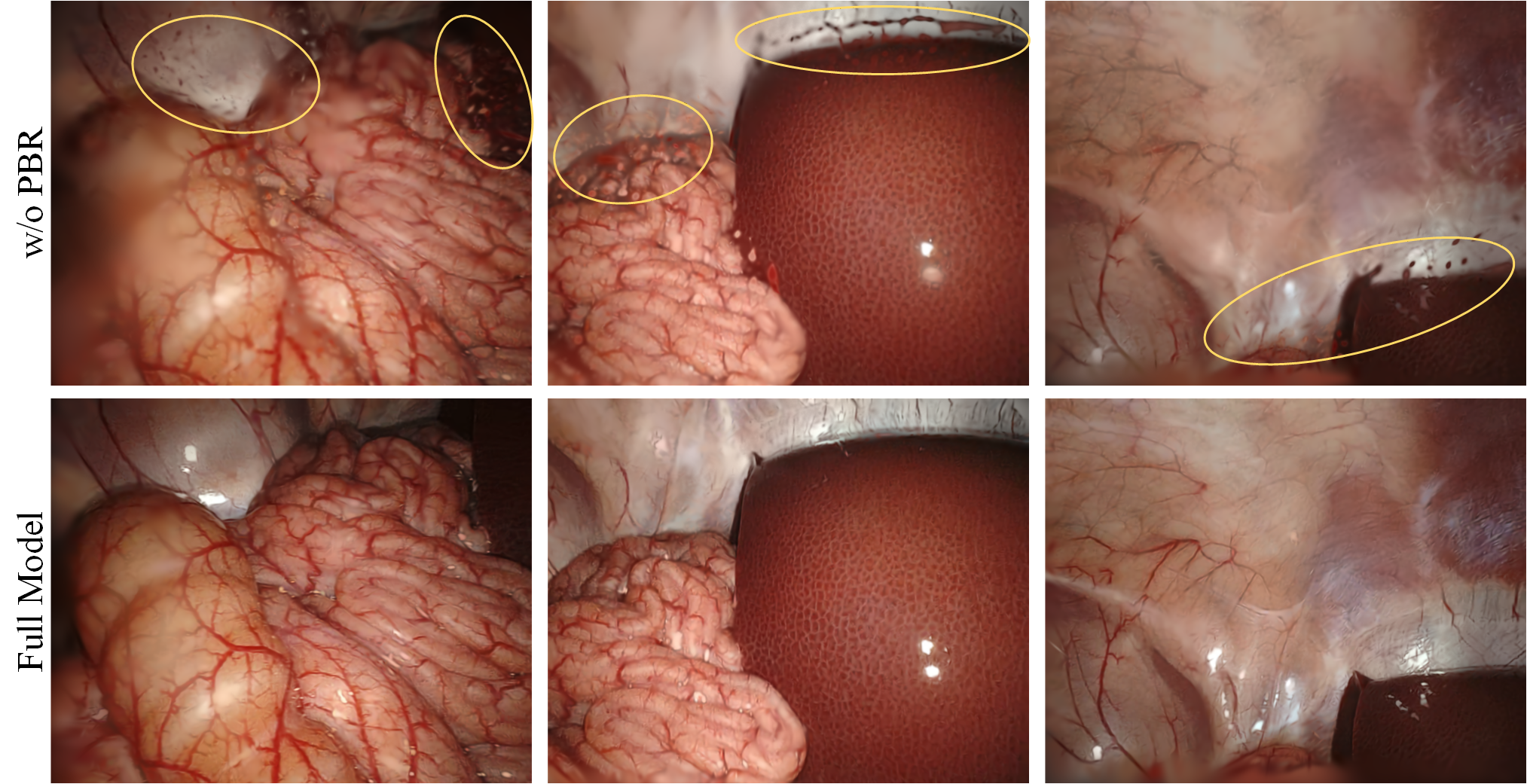}}
    % \vspace{-4pt}
\caption{\color{black}{Visual ablation study on the impact of Physics-Based Regularization (PBR). The top row shows the rendering results of the model without PBR (\textbf{w/o PBR}), while the bottom row presents the outputs of the \textbf{full model}. Regions highlighted in yellow indicate areas where the absence of PBR leads to visual artifacts, demonstrating that PBR helps preserve structural consistency and anatomical fidelity.}}
    % \vspace{-15pt}
    \label{fig:pbr_qualitative}
\end{figure}

\textbf{Loss Functions (LF).}
We introduce 2D Tracking Loss and Physics-Based Regularization to enhance our algorithm. To assess their effectiveness, we conduct three experiments: the \underline{full model}, the model without 2D Tracking Loss (\underline{w/o TL}), and the model without Physics-Based Regularization (\underline{w/o PBR}). The results in Table~\ref{tab:ablation} show that removing either TL or PBR reduces performance. This indicates that both losses provide useful constraints that improve reconstruction quality. {\color{black}In addition, the visual comparisons presented in Fig.~\ref{fig:pbr_qualitative} further demonstrate the effectiveness of the Physics-Based Regularization. As shown in the figure, the model trained without PBR (\underline{w/o PBR}) exhibits noticeable artifacts such as floating noise (highlighted in yellow). In contrast, the \underline{full model}, guided by physical priors, produces sharper details and more anatomically consistent reconstructions.}

\subsection{Effect of Monocular Depth Supervision}\label{sec:mono}

We investigate how using monocular depth maps as a supervision signal affects the performance of our method. As described in \eqref{equ:depth1} and \eqref{equ:depth2}, we obtain the aligned depth map \( D_{{fine}} \). Following previous work, we add an L1 depth loss during training. We perform experiments on Sequence2 of the StereoMIS dataset and report quantitative results in Table~\ref{tab:depth_supervision_ablation}. Our results show that monocular depth supervision achieves nearly the same performance as training without it, but slightly increases the training time (from 8.46 to 9.92 minutes). This suggests that, at present, monocular depth supervision does not clearly improve performance. We believe this is because the Depth Anything model \cite{yang2024depth} is trained on natural scenes, which limits its ability to generalize to surgical environments.

 \begin{table}[t]
\caption{Effect of Monocular Depth Supervision}
\label{tab:depth_supervision_ablation}
\centering
\setlength{\tabcolsep}{1.1pt}
\resizebox{\columnwidth}{!}{
\begin{tabularx}{\columnwidth}{l
>{\centering\arraybackslash}X 
>{\centering\arraybackslash}X 
>{\centering\arraybackslash}X 
>{\centering\arraybackslash}X}
\hline
Method 
& \cellcolor{blue!10}PSNR 
& \cellcolor{red!10}Abs Rel 
& \cellcolor{red!10}RMSE
& \cellcolor{red!10}Train Time \\
\hline
w/o Depth Sup.  & 31.49 & 0.129 & 9.271 & 8.46 min \\
w/  Depth Sup.  & 31.41 & 0.128 & 9.293 & 9.92 min \\
\hline
\end{tabularx}
}
% \vspace{-10pt}
\end{table}
\subsection{{\color{black}Hyperparameters}}

{\color{black}To assess the impact of varying loss component weights on the performance of Local-EndoGS, we trained the proposed framework using multiple weight configurations on the StereoMIS dataset. The quantitative results are illustrated in Fig.~\ref{fig:hyperparameter}.

\textbf{Photometric Loss Weight ($\lambda_{rgb}$):} 
As evidenced by the results, increasing $\lambda_{rgb}$ from $0.01$ to $1.00$ yields a continuous improvement in rendering quality, characterized by higher PSNR and lower RMSE values. This indicates that photometric consistency is fundamental to reconstruction quality, with higher weights encouraging the model to better capture fine image details. Consequently, we set $\lambda_{rgb}=1.0$ to ensure optimal rendering performance.

\textbf{Tracking Loss Weight ($\lambda_{track}$):} 
The experiments demonstrate that the model achieves the best trade-off between PSNR and RMSE at $\lambda_{track}=0.010$. A weight that is too small ($0.001$) fails to provide sufficient geometric guidance, whereas an excessively large weight ($0.100$) makes the optimization overly sensitive to noise in the 2D trajectory prior. This leads to geometric distortions and, consequently, degrades rendering quality.

\begin{figure*}[t]
\centerline{\includegraphics[width=\linewidth]{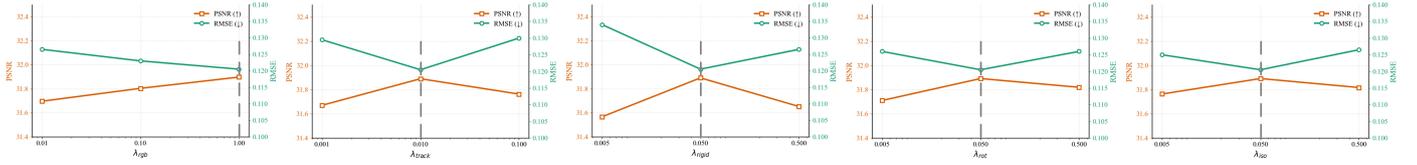}}
    % \vspace{-4pt}
\caption{Illustrations of the performance comparison with different hyperparameter configurations on the StereoMIS dataset.}
    % \vspace{-15pt}
    \label{fig:hyperparameter}
\end{figure*}

\textbf{Physical Regularization Weights ($\lambda_{rigid}, \lambda_{rot}, \lambda_{iso}$):} 
The rigidity, rotation, and isometry regularization terms exhibit highly consistent trends. The performance curves display distinct peaks (for PSNR) or troughs (for RMSE) in the central region of the parameter space.
At lower weights ($0.005$), the physical constraints are insufficient, resulting in non-physical artifacts within the deformation field.
At higher weights ($0.500$), excessive smoothing constraints restrict the flexibility of the Gaussian primitives, leading to a loss of high-frequency details.
The experiments identify $0.050$ as the optimal weight value. At this setting, the model maintains physical plausibility without compromising image reconstruction quality.
}

\section{Conclusion}
In this work, we present Local-EndoGS, a high-quality 4D reconstruction framework for deformable surgical scenes from monocular endoscopic sequences with arbitrary camera movements. Our approach combines a progressive window-based global representation, a local deformable scene representation, and a robust coarse-to-fine initialization strategy to effectively model complex tissue deformations and large camera motions. We also integrate long-range 2D pixel trajectory constraints and physical motion priors to improve the accuracy and physical validity of the reconstructed scenes. Extensive experiments on multiple datasets, including EndoNeRF, StereoMIS, and EndoMapper, show that Local-EndoGS achieves better performance than existing methods. Our framework has the potential to support various medical applications, including surgical planning and clinical training.

\textbf{Limitations and future work.}
Our method has several limitations that should be addressed in future research. First, because our framework is based on 3D Gaussian Splatting (3DGS), it inherits the limitation that 3D Gaussians cannot accurately represent surfaces due to multi-view inconsistency~\cite{huang20242d}. This restricts the accuracy of geometric reconstruction and the recovery of fine details. In the future, we will explore ways to improve the multi-view consistency of 3D representations, such as integrating advanced surface regularization techniques~\cite{guedon2024sugar} or developing hybrid representations that combine the strengths of Gaussians and implicit neural fields~\cite{yu2024gsdf}.
Second, like most existing methods for reconstructing deformable surgical scenes~\cite{yang2024deform3dgs,eh-surgs,li2024endosparse,xie2024surgicalgaussian,liu2024lgs,huang2024endo,liu2025foundation}, our approach is designed for offline reconstruction. It produces high-quality deformable reconstructions that are suitable for treatment planning, surgical education, and dataset creation. However, it is not suitable for real-time applications. In future work, we will focus on improving computational efficiency and optimizing our algorithms to enable real-time deformable reconstruction for intraoperative surgical use. 
{\color{black}Moreover, our current framework incorporates physical priors, including a local isometry prior that assumes continuous tissue deformation without topological change and constrains surface patches to preserve local geometry, which may not fully capture events such as cutting or tearing. Extending the model to handle topological changes (e.g., through adaptive surface representations or an event detection module) will be an important direction for future research.}
In addition, another limitation concerns our training strategy. Specifically, we first divide the scene into local windows and then train each window sequentially. Although this approach allows us to handle long sequences with arbitrary camera motion and improves reconstruction performance (as shown in Sec.~\ref{sec:ablation}), the total training time increases linearly with the number of windows. Moreover, this sequential process does not fully utilize the parallel processing capabilities of modern GPUs, leading to suboptimal computational efficiency. In the future, we will develop parallel training strategies to process multiple local windows simultaneously and design more refined inter-window consistency mechanisms. This will reduce overall training time and make better use of GPU resources. 
{\color{black}Finally, incorporating specialized optical-flow or multi-view 3D matching models (e.g., MFT \cite{neoral2024mft}, MASt3R \cite{leroy2024grounding}) into our initialization framework represents another promising direction for future work. Exploring the establishment of correspondences through these methods and leveraging their respective strengths could further improve the quality and robustness of our reconstruction framework.}
		
\section{Declaration of Competing Interest}
		The authors declare that they have no known competing financial interests or personal relationships that could have appeared to influence the work reported in this paper.

	\section{Acknowledgement}
This work was supported in part by the National Key R\&D Program of China (Grant No.2023YFB4705700), in part by the Natural Science Foundation of China under Grant 62225309, U24A20278, 62361166632, U21A20480, 6240331 and 62203298, in part by State Key Laboratory of Robotics and Intelligent Systems (No: 2024-O26), in part by Innovation and Technology Commission of Hong Kong (ITS/235/22, ITS/225/23, ITS/224/23, MHP/096/22 and Multi-scale Medical Robotics Center, InnoHK), and in part by Research Grants Council of Hong Kong (CUHK 14217822, CUHK 14207823, CUHK 14211425, T45-401/22-N and AoE/E-407/24-N).

	\bibliographystyle{model2-names.bst}\biboptions{authoryear}
	\bibliography{refers}
\end{document}